\documentclass[acmtog,authorversion]{acmart}

\usepackage{nicefrac}
\usepackage{colortbl}  
\newcommand{\TABLERED}{\cellcolor[rgb]{0.92,0.67,0.67}}
\newcommand{\TABLEGREEN}{\cellcolor[rgb]{0.76,0.98,0.70}}
\newcommand{\TABLEYELLOW}{\cellcolor[rgb]{0.98,0.99,0.70}}
\renewcommand\footnotetextcopyrightpermission[1]{} 
\begin{document}

\title{%
Centimeter-Wave Free-Space Time-of-Flight Imaging
} 


\author{Seung-Hwan Baek*, Noah Walsh*, Ilya Chugunov, Zheng Shi, \lowercase{and} Felix Heide}
\affiliation{%
  \\\institution{Princeton University}
}
\renewcommand{\shortauthors}{Baek,Walsh,Chugunov,Shi,Heide}

\thanks{*S.~H. Baek and N. Walsh contributed equally to this work.}

\begin{abstract}
Depth cameras are emerging as a cornerstone sensor modality with diverse application domains that directly or indirectly rely on robustly and accurately measured scene depth, including personal handheld devices, robotics, scientific imaging, and self-driving vehicles. Although time-of-flight (ToF) depth sensing methods have fueled these applications, the precision and robustness of pulsed and correlation ToF methods is fundamentally limited by relying on photon time-tagging or modulation \emph{after} photo-conversion. Successful optical modulation approaches have been restricted fiber-coupled modulation with large coupling losses or interferometric modulation with sub-cm range, and the {precision} gap between interferometric methods and ToF methods is more than three orders of magnitudes. In this work, we close this gap and propose a computational imaging method for \emph{all-optical free-space correlation before photo-conversion} that achieves micron-scale depth resolution with robustness to surface reflectance and ambient light with conventional silicon intensity sensors and continuous-wave laser illumination. To this end, we solve two technical challenges: modulating at GHz rates and computational phase unwrapping. We propose an imaging approach with resonant polarization modulators and devise a novel \emph{optical dual-pass frequency-doubling} which achieves high modulation contrast at more than 10~GHz -- two orders of magnitude higher than the analog modulation in silicon correlation ToF sensors. At the same time, \emph{centimeter-wave modulation together with a small modulation bandwidth render existing phase unwrapping methods ineffective.} We tackle this problem with a data-driven \emph{neural phase unwrapping method}, inspired by modern segmentation networks, that exploits that adjacent wraps are often highly correlated. We validate the proposed method in simulation and experimentally, where it achieves micron-scale depth precision. We demonstrate precise depth sensing independently of surface texture and ambient light and compare against existing analog demodulation methods, which we outperform across \emph{all} tested scenarios.
\end{abstract}

\begin{teaserfigure}
 \centering
  \includegraphics[width=\linewidth]{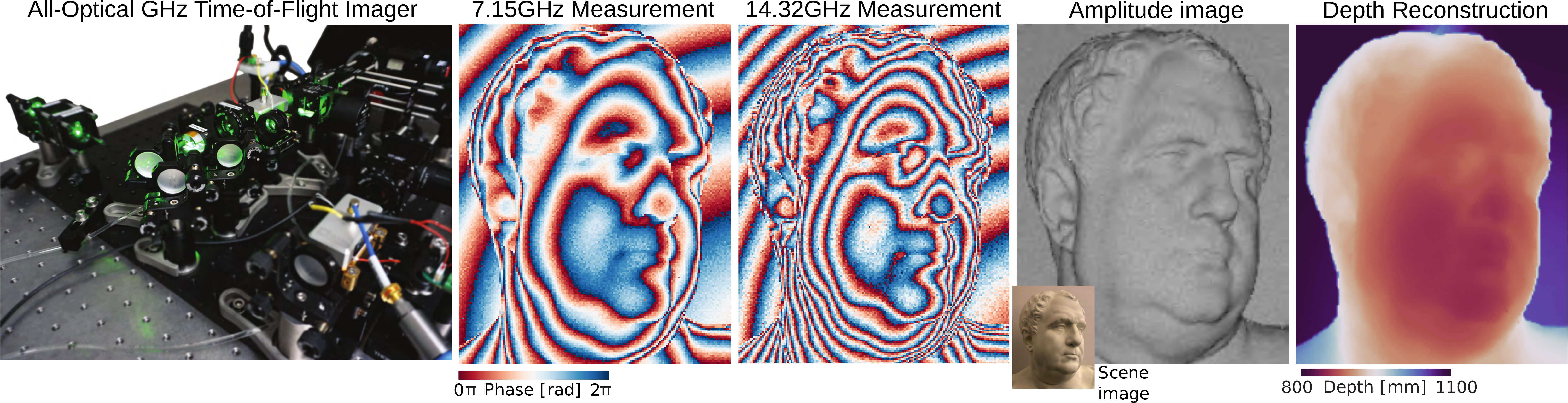}%
  \caption[]{\label{fig:teaser}%
   We propose an all-optical time-of-flight imager operating at GHz modulation frequencies. To modulate light without optical couplers, we introduce a novel dual-pass modulation scheme using polarization optics combined with a custom electro-optical resonant polarization modulator. The proposed system allows us to capture correlation signals that are optically computed in free-space before photo-conversion -- without the need of analog modulation on the sensor or optical coupling. To unwrap the centimeter-wave correlation images, we devise a novel neural unwrapping method that poses high-frequency time-of-flight phase unwrapping as an ordinal classification problem. We demonstrate high-precision depth imaging for macroscopic scenes from a single- and double-frequency measurement pair.
   }
\end{teaserfigure}

\maketitle

\section{Introduction}
\label{sec:introduction}

Cameras that acquire scene depth alongside conventional color imagery have emerged as an enabling imaging modality with broad applications across domains, ranging from 3D scanning in personal devices such as the lidar-equipped Apple iPhone 12, interactive gaming using Microsoft's Kinect One, virtual and augmented reality, precise industrial and robotic scanning in close ranges, to autonomous vehicles and delivery drones that must see and understand scenes at large standoff distances.
RGB-D depth cameras have made it possible to recover high-fidelity scene reconstruction~\cite{izadi2011kinectfusion,tulsiani2018multi}, which effectively drive 3D scene understanding~\cite{song2015sun,hickson2014efficient} and 3D deep learning~\cite{dai2018scancomplete} with the collection of large 3D datasets \cite{silberman2012indoor, chang2015shapenet, dai2017scannet} as the most successful 3D learning methods are supervised today. As such, emerging depth camera technologies does not only \emph{directly} drive application fields but also \emph{indirectly} fuel modern 3D deep learning and scene understanding.

The most successful depth acquisition approaches for wide operating ranges are based on active time-of-flight (ToF) sensing that allows for high depth precision at small sensor-illumination baselines~\cite{hansard2012time}. Passive approaches that infer distances solely from monocular camera images~\cite{subbarao1994depth, mahjourian2018unsupervised} or from parallax of binocular camera pairs~\cite{hirschmuller2005accurate, kendall2017end, xu2020aanet} do not offer the same range and depth precision as they struggle with textureless regions and complex geometries~\cite{smolyanskiy2018importance,lazaros2008review}. Active sensing approaches tackle this challenge by emitting light into the scene and reconstructing depth from the returned signal. Structured light methods such as active stereo systems use spatially patterned light to aid estimating depth from stereo matching~\cite{ahuja1993active}. While being robust to textureless scenes, their accuracy is fundamentally limited by illumination pattern density and sensor baseline, resulting in a large camera form-factor. ToF depth sensing approaches lift these limitations by estimating depth via the travel time of photons leaving from and returning to the device, allowing for co-axial sensor setups with no illumination-camera baseline.

Direct ToF sensor systems, such as lidar~\cite{schwarz2010lidar}, acquire point clouds directly from individual pulses sent into a scene. These systems \emph{directly} measure the round-trip time of emitted pulses of light. However, this direct acquisition approach is fundamentally limited by the amount of signal photons present in a reflected pulse, decreasing quadratically with distance for diffuse objects. Increasing the energy of the illumination pulse is not a viable alternative as high peak power ultra-short pulses cannot be realized in a small form-factor and are fundamentally limited by eye-safe levels~\cite{schwarz2010lidar,spinhirne1995compact,williams_apds}. \emph{Direct acquisition} also mandates sensitive time-resolved detectors such as single-photon avalance diodes (SPADs)~\cite{aull2002geiger,bronzi2016automotive,niclass2005design,rochas2003first} that suffer from low photon-detection efficiency in CMOS technology~\cite{villa2014cmos}. As a result, directly capturing the temporal impulse response fundamentally limits the measurement signal to noise ratio (SNR) by the detection and time-tagging of the few photons in an ultra-short pulse that indirectly return to a detector. Hence, the most successful direct ToF cannot acquire precise depth estimates without averaging \emph{millions} of picosecond pulses~\cite{heide2018sub}.

Correlation time-of-flight methods~\cite{heide2013low, su2018deep, heide2015doppler, shrestha2016computational, heide2014imaging} attempt to lift these limitations by \emph{indirectly} recovering the travel time of light. These cameras flood a scene with periodic amplitude-modulated light, and calculate the phase shift of returned light to infer depth. In contrast to direct ToF sensing approaches, the continuous intensity modulation does not require ultra-short pulse generation and time-tagging of individual photons in a pulse. Correlation ToF sensors that demodulate the amplitude-modulated flash-illumination on-sensor have been widely adopted, such as in the Microsoft Kinect One camera. These sensors implement multiple charge buckets per pixel and shift a photo-electron to an individual bucket by applying an electrical potential between the individual quantum wells~\cite{lange2001solid}. While amplitude modulation allows for depth precision {comparable} to picosecond-pulsed direct ToF, and three orders of magnitude lower cost thanks to the continuous wave modulation and CMOS sensor implementation, it is also this sensing mode which fundamentally limits such depth sensors. Specifically, the \emph{modulation after photo-electric conversion limits the maximum achievable modulation frequency to 200~MHz}, restricted by the photon absorption depth in silicon~\cite{lange2001solid}. As such, the depth resolution of these sensors is fundamentally limited by the maximum modulation frequency $\omega_\text{max}$, corresponding to a resolution of $\nicefrac{c}{(360\cdot \omega_\text{max}) }$ per $1^\circ$ phase~\cite{lange2001solid}, and, hence, existing CMOS multi-bucket sensors are restricted to achieve sub-cm-precision. At the same time, low modulation contrast due to the on-sensor analog modulation at at high frequencies makes existing \emph{correlation ToF imagers struggle for textured objects with low reflectance or highly specular objects with weak diffuse reflectance.} Approaches that propose fiber-coupled modulation techniques from optical communication~\cite{kadambi_rethinking} to observe beat-patterns as an alternative struggle with high coupling loss and low modulation contrast. Specifically, \emph{coupled grating modulation results in coupling losses of more than 100~$\times$} in recent designs, including ~\cite{rogers2021universal,marchetti2017high,bandyopadhyay2020highly}, restricting these approaches to theoretical works with existing fiber-optic modulators. This also applies to the very recent work from Rogers et al.~\cite{rogers2021universal}, which the approach proposed in this work outperforms by an order of magnitude in depth precision.

Gupta et al.~\shortcite{coding_functions_gupta} have reported the most successful correlation time-of-flight method beyond 100~MHz modulation bandwidth. By employing analog radio-frequency (RF) modulation \emph{after photo-conversion} their system achieves a modulation bandwidth of up to 500~MHz allowing them to include high-frequency components in their modulation scheme, achieving mm-resolution. The only depth acquisition approaches that have overcome this modulation bandwidth rely on \emph{optical interferometric modulation before photo-conversion}. However, successful optical modulation approaches have been limited to interferometric modulation, which achieve micron-resolution but are also limited to cm ranges~\cite{gkioulekas2015micron,li2018SH-ToF} impractical for large scenes. As such, the resolution gap between interferometric methods and the most successful correlation ToF methods is more than three orders of magnitudes.

In this work, we close this gap and depart from existing approaches that modulate after photo-conversion. We propose a computational imaging method with a novel all-optical dual-frequency modulation approach in free-space that achieves micron-scale depth resolution with continuous-wave laser illumination. To this end, \emph{we lift two fundamental limitations of existing approaches: coupler-free amplitude modulation at GHz rates and phase unwrapping with cm-wavelengths.} To modulate intensity of light in freespace at such high rates, we propose a polarization-modulation approach using custom resonant electro-optic modulators (EOM) to free-space modulate the incident and emitted illumination. This approach allows us achieve $> 50\%$ modulation contrast for large-area freespace modulation at 7~GHz with a few volts, and we devise an optical method that allows this \emph{frequency to be optically doubled to 14~GHz by passing twice through the same modulator}. While the proposed resonance operation allows for a high fundamental frequency, it also limits the bandwidth to a few MHz. As such, without lower frequency bands available, GHz-frequency time-of-flight imaging produces extensive phase wrapping. For example, the modulation frequency of 10~GHz amounts to 1.5~cm physical range within a single phase wrap, and a target at a meter distance corresponds to a phase offset of 67 wraps, in contrast to MHz correlation ToF cameras with a wavelength of several meters corresponding to single-digit wraps. Unfortunately, existing micro-frequency ToF unwrapping~\cite{gupta2015phasor} also does not offer an alternative due to the limited modulation bandwidth that renders look-up table-based approaches ineffective in the presence of measurement noise. We solve this problem using a novel data-driven \emph{neural phase unwrapping method and tackle this challenge as a learned classification problem}, where each pixel of the measured data is mapped to a wrap count. We use a trained convolutional neural network to map the predictions into the correct un-aliased bins and remove these erroneous wrapping artifacts, trained in an end-to-end fashion on simulated time-of-flight data generated from a diverse set of ground truth depth scenes. Our system thus does \emph{not follow Skellam-Gaussian noise statistics as conventional AMCW ToF correlation sensors do, but instead Poisson-Gaussian noise} which we model in simulation to train a noise-robust phase unwrapping network. In contrast to semantic segmentation, the proposed network exploits that adjacent classes are often highly correlated.

We validate the proposed computational imaging method in simulation and experimentally, demonstrating a resolution below 30~micron over macroscopic room-sized distances at mW laser powers, corresponding to < 100 femtosecond temporal resolution. Jointly with the learned unwrapping, the all-optical modulation without coupling losses, allows for robustness to low-reflectance texture regions and highly specular objects with low diffuse reflectance component. We assess the neural phase unwrapping network extensively in simulation and experimentally, validating that it outperforms existing conventional and learned unwrapping approaches across all tested scenarios. We further validate precision and compare extensively against post-photoconversion modulation, which fail in low flux scenarios, and interferometric approaches, that are limited to small ranges. As our free-space modulation is all-optical, it allows we demonstrate that it can be readily combined with interferometric modulation, allowing us to close the gap between both --- achieving micron precision over macroscopic scenes, and picosecond temporal sampling, with the potential for fueling photon-efficient imaging of ultrafast phenomena in chemistry, biology or physics.

Specifically, we make the following contributions in this work:
\begin{itemize}
	\setlength\itemsep{.5em}
	\item We present a computational time-of-flight imaging approach with fully optical free-space correlation that allows for $\geq$~10 GHz frequencies.
	\item As part of this method, we introduce a novel neural unwrapping and denoising method that poses high-frequency time-of-flight phase unwrapping as a classification problem.
	\item We introduce a novel two-pass optical frequency doubling approach for intensity modulation using resonant electro-optical modulators.
	\item We validate the proposed method experimentally with a prototype system, demonstrating micron-scale depth resolution for macroscopic scenes and depth estimates robust to surface reflectance and ambient light.
\end{itemize}
To ensure reproducibility, we will share the schematics, code, and optical design of the proposed method.

\vspace{5pt}
\paragraph{Overview of Limitations and Reproducibility}
While the proposed amplitude modulation approach is an all-optical free-space modulation, the proposed modulators currently support an active area of 2.5 $\times$ 2.5~mm. While a telescope optical system could allow for 2D modulation with the proposed modulators, then also allowing us to use 2D sensor arrays, we left this engineering effort as out of scope for this work. In the future, we plan to have modulators with larger optical area fabricated to facilitate full-sensor modulation. As a result, of the galvo scanning used in our setup and others~\cite{coding_functions_gupta,li2018SH-ToF}, the proposed method requires sequential scanning as in lidar sensing~\cite{schwarz2010lidar}.

\section{Related Work}
\label{sec:relatedwork}
In this section we seek to give the reader a broad overview of the current state of depth imaging methods, in order to better describe the gap our work fills in the 3D vision ecosystem. A qualitative summary of depth acquisition approaches discussed below is also given in Table~\ref{tab:summary}.

\paragraph{Depth Imaging}
There exists a wide family of depth imaging methods in use today. These can be broadly categorized into passive methods, which leverage solely image cues such as parallax \cite{hirschmuller2005accurate,baek2016birefractive,meuleman2020single} and defocus \cite{subbarao1994depth} to infer depth, and active methods, which first project a known signal into the scene before attempting to recover depth. While passive approaches can offer low-cost depth estimation methods, using commodity camera hardware~\cite{garg2019learning}, they struggle to achieve sub-cm accuracy in best case scenarios, and can fail catastrophically for complex scene geometries and textureless regions \cite{smolyanskiy2018importance}. Structured light approaches, such as those used in the Kinect V1, combat these failure cases by improving local image contrast with active illumination~\cite{scharstein2003high, ahuja1993active}, at a detriment to form-factor and power consumption. These methods, however, still cannot disambiguate mm-scale features, as they are smaller than illumination feature size itself and so stereo correspondences cannot be accurately found. ToF imaging is an active method that does not rely on visual cues, avoiding the pitfalls of stereo matching completely. ToF cameras instead directly or indirectly measure the travel time of light to infer distances~\cite{lange2001solid, hansard2012time}, with modern continuous-wave correlation ToF systems achieving mm-level accuracy for megahertz-scale modulation frequencies.
Interferometry extends this principle to the terahertz range, exploiting the interference of electromagnetic waves to measure their travel time. These systems can achieve micron-scale accuracy at the cost of mm-scale operating ranges~\cite{hariharan2003optical}. In this work we seek to bridge the gap between commodity megahertz frequency ToF systems and terahertz frequency interferometry; obtaining micron-scale depth resolution beyond cm-scale scenes.

\paragraph{Pulsed ToF}
Pulsed ToF systems are direct ToF acquisition methods. These methods send discrete laser pulses into the scene and measure their time-to-return via avalanche photodiodes~\cite{cova1996avalanche,pandey2011ford} or single-photon detectors~\cite{mccarthy2009long,heide2018sub,gupta2019photon,gupta2019asynchronous}, which can offer improved sensitivity and operating range. This ability can allow pulsed ToF methods to image over hundreds of meters with relatively high SNR, however this time-resolved approach means their accuracy is limited by the temporal resolution of their sensors. As these typically operate with hundreds of picosecond-scale quantization, their precision is also typically limited to cm for indoor distances and 10s of cm for large outdoor scenes~\cite{behroozpour2017lidar}. Peak pulse power is also limited by eye safety constraints, meaning that while short pulse widths are desired to improve depth resolution, the resulting system must operate with increasingly sparse photon counts~\cite{heide2018sub}, requiring \emph{millions} of picosecond pulses~\cite{heide2018sub} for mm-scale resolution. In this work, we revisit indirect ToF using amplitude modulation as an alternative to achieve sub-picosecond resolution without time-resolved sensors and time-tagging electronics but conventional intensity sensing.
\begin{table}[t]
	\centering
	\resizebox{\columnwidth}{!}{
	\begin{tabular}{c|c|c|c|c|c}
		\hline
		Method & Depth & Operating & Photon  & System & Reflectance\\
		 & Resolution & Range & Efficiency & Cost & Robustness\\
		\hline \hline
    Pulsed TOF& \TABLEYELLOW cm & \TABLEGREEN $>$ m & \TABLERED low & \TABLERED high & \TABLEYELLOW medium\\ \hline
		Stereo & \TABLEYELLOW cm & \TABLEYELLOW m & \TABLEYELLOW medium & \TABLEGREEN low & \TABLERED low\\ \hline
		FMCW ToF & \TABLEYELLOW cm  & \TABLEYELLOW m & \TABLEYELLOW medium & \TABLEYELLOW medium & \TABLEGREEN high \\ \hline
		Interferometry & \TABLEGREEN $\mu$m & \TABLERED mm & \TABLEGREEN high & \TABLEYELLOW medium &\TABLEYELLOW medium \\ \hline
		MHz AMCW ToF & \TABLEYELLOW cm  & \TABLEYELLOW m & \TABLEYELLOW medium & \TABLEGREEN low & \TABLEYELLOW medium \\ \hline
    GHz Heterodyning & \TABLEGREEN $\mu$m & \TABLEYELLOW m & \TABLERED low & \TABLEGREEN low & \TABLEYELLOW medium \\ \hline
		Ours & \TABLEGREEN $\mu$m & \TABLEGREEN $>$ m & \TABLEGREEN high & \TABLEGREEN low & \TABLEGREEN high \\
		\hline
	\end{tabular}
	}
	\caption{	\label{tab:summary}%
Comparison of depth imaging methods for depth resolution, operating range, photon efficiency, robustness to texture and reflectance, and cost. This table provides a categorization of the space of technologies and exact numbers may vary for different device configurations.
}
\vspace{-6mm}
\end{table}

\paragraph{Interferometry and Frequency-Modulated Continuous-Wave ToF}
Optical interferometry methods leverage the interference of electromagnetic waves to infer their path lengths, information which is encoded in the measured amplitude and/or phase patterns. A detailed review of interferometry can be found in \cite{hariharan2003optical}. Methods such as optical coherence tomography (OCT)~\cite{huang1991optical} have found great use biomedical applications~\cite{fujimoto2016development} for their ability to resolve micron-scale features in optical scattering media. This, however, comes with the caveat of a mm-scale operating range as diffuse scattering leads to a sharp decline in SNR. In graphics, OCT approaches have been successfully employed to achieve micron-scale light transport decompositions~\cite{gkioulekas2015micron} and light transport probing~\cite{kotwal2020interferometric}.
{
Fourier-domain OCT systems mitigate some of the sensitivity to vibration by using a spectrometer and a broadband light source~\cite{leitgeb2003performance}.
}
While these methods provide high temporal resolution, they are also limited to cm-scale scenes.
Frequency-modulated continuous-wave (FMCW) ToF systems employ an alternative interferometric approach to measuring distance. These methods continuously apply frequency modulation to their output illumination, which when combined in a wave-guide with the delayed returned light from the scene produces constructive and destructive interference patterns from which travel-time (and thereby depth) can be inferred. Experimental CMOS-based FMCW LiDAR setups can achieve millimeter precision for scenes at decimeter range~\cite{behroozpour2016electronic}, but require complex tunable laser systems that can be susceptible to phase noise unless locked to operating within a thin frequency band~\cite{sandborn2016dual, amann1992phase}.
We propose continuous-wave intensity modulation, allowing us to use conventional continuous wave lasers modulated and demodulated in free-space.

\paragraph{Amplitude-Modulated Continuous-Wave ToF}
Amplitude-modulated continuous wave (AMCW) ToF sensors flood the scene with periodically modulated light and infer distance based on accrued phase differences in the returned light~\cite{lange2001solid,hagebeuker20073d,remondino2013tof}. These systems, such as cameras in the prolific Microsoft Kinect series~\cite{zhang2012microsoft}, can rely on affordable CMOS sensors and conventional CW laser diodes to produce dense depth measurements. A large body of work explores methods to mitigate multipath-interference~\cite{achar2017epipolar,fuchs2010multipath,freedman2014sra,kirmani2013spumic,bhandari2014resolving,multipath_kadambi,jimenez2014modeling,naik_light_transport_model} or use correlation ToF measurements to resolve travel-time of light in flight~\cite{heide2013low,multipath_kadambi}. The time-resolved transient images captured by these methods have found a number of emerging applications, such as non-line-of-sight imaging~\cite{heide2014imaging,occluded_imaging_tof}, imaging through scattering media~\cite{heide2014imaging}, or material classification~\cite{su2016material}. While this body of work illustrates the potential of correlation ToF imaging, they are, however, restricted to operating at $\approx$100MHz modulation frequencies due to photon absorption depth in silicon~\cite{lange2001solid}, which itself governs how these devices perform photo-electric conversion. This limit places the depth resolution of modern AMCW ToF sensors at mm-scale for operating ranges of up to several meters. Works such as \shortcite{kadambi2017rethinking, li2018SH-ToF} attempt to circumvent this limit and push AMCW ToF cameras to GHz modulation frequencies.
However, fiber coupling reduces light efficiency by a factor of more than 100$\times$ compared to the proposed method~\cite{marchetti2017high,bandyopadhyay2020highly}, drastically reducing precision and range with eye-safe laser power levels. This also applies to the very recent work from Rogers et al.~\shortcite{rogers2021universal}, which this work outperforms by an order of magnitude in precision. The approach from Li et al.~\shortcite{li2018SH-ToF} overcomes some of these limitations but relies on interferometric modulation only making the method susceptible to speckle, vibration, laser frequency drift and other common interferometry errors. The most successful method for AMCW ToF beyond 150~MHz bandwidth from Gupta et al.~\shortcite{coding_functions_gupta} and Gutierrez et al.~\shortcite{gutierrez2019practical} employs a fast photodiode and analog radio-frequency (RF) modulation \emph{after photo-conversion} allowing to include high-frequency components in their modulation scheme with mm-resolution. In this work we move this correlation to the optical domain \emph{in free-space, before photo-conversion or fiber-coupling}, and, when combined with computational phase unwrapping, this allows us to with which we achieve micron-scale depth resolution, resilient to ambient light and sources of error that plague existing GHz frequency setups.

\begin{figure*}[t]
    \centering
    \includegraphics[width=\linewidth]{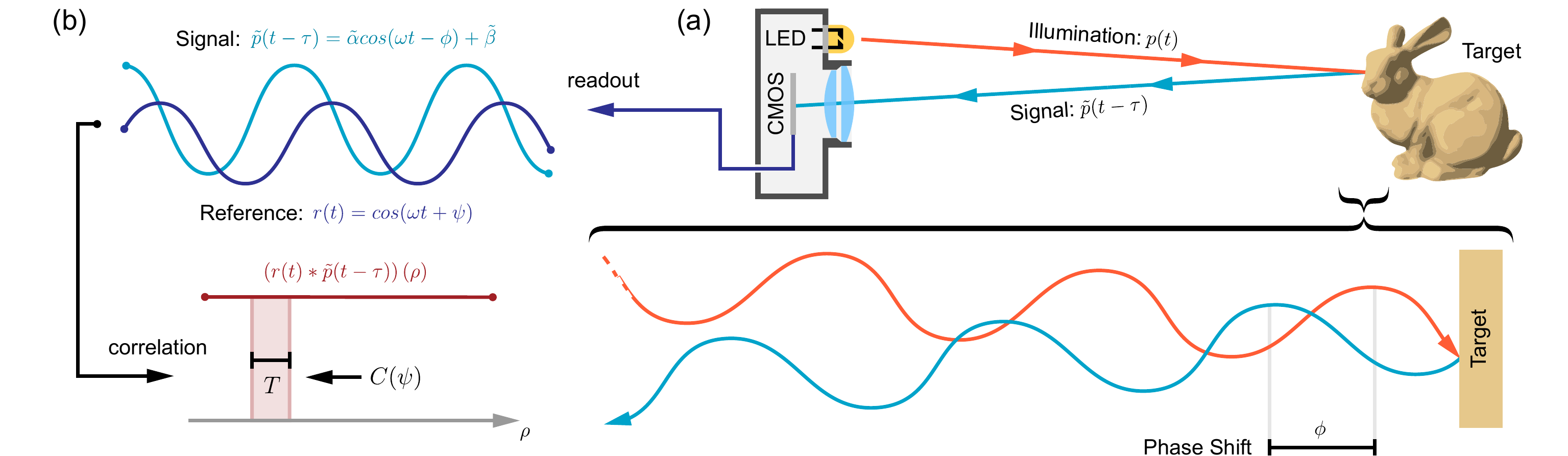}
    \caption{\label{fig:tof_imaging}%
    Principle of AMCW ToF homodyning.
    (a) Typical AMCW ToF imagers emit coded illumination into a scene with a time-varying sinusoidal intensity modulation. The reflected light encodes travel time via the phase shift of the returned modulated signal.
    (b) Homodyne detectors measure the correlation between the reflected sinusoidal signal and a reference signal with same frequency, resulting in a DC value as a function of the reference-signal phase and the phase shift from the scene.
    }
\end{figure*}

\paragraph{Depth Reconstruction and Phase Unwrapping}
Depth cameras are susceptible to a wide scope of noise, including errors stemming from object reflectance properties, and signal interference like ambient light. This has sparked a large body of work in both denoising algorithms \cite{lenzen2013denoising, yan2018ddrnet, sterzentsenko2019self} which seek to remove this noise from depth maps in post-processing, as well as end-to-end depth reconstruction approaches which attempt to directly output high-fidelity 3D data from raw measurements \cite{su2018deep, qiu2019deep, marco2017deeptof, Guo_2018_ECCV}.
In AMCW ToF systems the analog correlation signal is prone to be phase-shifted by more than one wavelength. This means to recover the true phase offset, and thereby accurately reconstruct depth, phase unwrapping algorithms~\cite{dorrington2011separating,crabb2015fast,lawin2016efficient, an2016pixel} are required.  Single phase unwrapping approaches~\cite{crabb2015fast,ghiglia1998two, bioucas2007phase, bioucas2008absolute} are only able to recover the relative depth and wrap count, and require a-priori assumptions to estimate scale.  Multi-frequency phase unwrapping methods solve this limitation by unwrapping high-frequency phases with their lower-frequency counterpart.  Wrap count is recovered by either weighing Euclidean division candidates ~\cite{bioucas2009multi, droeschel2010multi, kirmani2013spumic, freedman2014sra, lawin2016efficient} , or using frequency-space lookup table ~\cite{gupta2015phasor}. All of these methods, while powerful for MHz ToF imaging, fail in the presence of noise for the the hundreds of wrap counts observed in the GHz correlation imaging. To tackle this challenge, in this work we propose a neural network capable of simultaneously denoising and unwrapping GHz frequency ToF correlation measurements.
\section{Correlation ToF Imaging}
\label{sec:image_formation}
In this section we provide an overview of the principles of amplitude modulated continuous wave (AMCW) ToF imaging. In the following, we assume a sine wave model for notational simplicity, although the proposed method is not limited to sinusoidal modulation.

\paragraph{Image Formation}
Correlation AMCW ToF cameras begin by emitting an amplitude-modulated light signal into the scene, that is
\begin{equation}\label{eq:light_source}
    p(t) := \alpha \cos(\omega_p t) + \beta,
  \end{equation}
where $\omega_p$ is modulation frequency,  $\alpha$ is amplitude, and $\beta$ is a DC offset. This light is then captured by the ToF camera after travel time $\tau$, which is dependent on the scene depth.
The measured signal $\tilde{p}$ is the time delayed signal of $p$ with changes in amplitude, phase, and bias as
\begin{equation}\label{eq:detected}
    \tilde{p}(t - \tau) = \tilde{\alpha} \cos(\omega_p t - \phi) + \tilde{\beta}, \quad \phi = 2\pi \omega_p \tau.
\end{equation}
We observe an attenuation in amplitude $\tilde{\alpha}$, a shift in bias $\tilde{\beta}$, and a $\tau$-dependent phase shift $\phi$. As illustrated in Figure~\ref{fig:tof_imaging} (a), this phase shift occurs due to the round trip distance of the illumination light not being a perfect integer multiple of its wavelength. Correlation ToF cameras correlate this measured signal with a reference $r(t)=\cos(\omega_r t + \psi) + 1/2$, where $\omega_r$ is the demodulation frequency and $\psi$ is the demodulation phase. In existing multi-bucket correlation imagers, this correlation occurs during exposure via photonic mixer device (PMD) pixels~\cite{lange2001solid,foix2011lock}, which are modulated according to the reference function $r(t)$. Expanding this correlation we see that
\begin{align}\label{eq:correlation_expanded}
    \tilde p( {t - \tau } )r( {t} ) \;=\; &\left( \tilde{\alpha} \cos(\omega_p t - \phi) + \tilde{\beta} \right) \left( \cos(\omega_r t + \psi) + \frac{1}{2} \right) \nonumber\\
    \;=\; &\frac{\tilde{\alpha}}{2}\cos((\omega_p - \omega_r) t - \phi + \psi)\; + \nonumber\\
     &\frac{\tilde{\alpha}}{2}\cos((\omega_p + \omega_r) t + \phi + \psi)\; + \nonumber\\
     &\tilde{\beta}\cos(\omega_r t + \psi) + K,
\end{align}
where $K$ is a general constant offset, meant to model the non-zero modulation on the sensor.
Given this correlation measurement, we aim to estimate the phase delay $\phi$ from which the scene depth can be computed.

\paragraph{Homodyne Modulation}
When the modulation and demodulation frequencies are the same ($\omega_p = \omega_r = \omega$) this is called \textit{homodyne} modulation. We refer the reader to the Supplemental Material for a detailed discussion of heterodyne modulation ($\omega_p \neq \omega_r$)~\cite{conroy2009range}. For homodyne modulation, Equation~\ref{eq:correlation_expanded} becomes
\begin{align}\label{eq:correlation_homodyne}
    \tilde p( {t - \tau } )r( {t} ) \;=\; &\frac{\tilde{\alpha}}{2}\cos(\psi - \phi) + K \;  +\nonumber\\
    &\frac{\tilde{\alpha}}{2}\cos(2 \omega t + \phi + \psi)\; + \tilde{\beta}\cos(\omega t + \psi).
\end{align}
After integration over exposure time $T$, we get a correlation measurement
\begin{equation}\label{eq:correlation}
    C_\psi(\rho) = \int_{\rho}^{\rho+T} {\tilde p( {t - \tau } )r( {t} ) \, \mathrm{d}t} = \frac{\tilde{\alpha}}{2}\cos(\psi - \phi) + TK. \\
\end{equation}
We note here that for $T \gg 1/\omega$, this integration acts a low-pass filter plus sampling, which eliminates the latter two high-frequency terms in Equation~\ref{eq:correlation_homodyne}. As illustrated in Figure~\ref{fig:tof_imaging} (b), $C_\psi(\rho)$ is a constant, varying in magnitude with the value of $\psi$ (achieving its maximum at $\psi=n\phi, n\in\mathbb{N}$). In practice, this means we never have to explicitly temporally sample $\tilde p(t-\tau)$, which would require expensive and photon-inefficient ultrafast detectors and RF modulation electronics. Although $C_\psi(\rho)$ does not directly give us access to the true value of $\phi$, by sampling this function at different demodulation phase offsets $\psi$ we can make use of Fourier analysis to discern $\phi$. Existing correlation imagers typically acquire four equally-spaced phase values $\psi_i\!=\!0,\, \pi/2,\, 3\pi/2,\, \pi$~\cite{lange2001solid}. Using these phase values we can estimate the measured signal's $\phi$ down to a $2\pi$ integer phase wrapping factor
\begin{equation}\label{eq:depth}
    \phi = \arctan\left(\frac{C_{\pi}(0)-C_{\pi/2}(0)}{C_{0}(0)-C_{3\pi/2}(0)}\right) + 2\pi n. \quad n \in  \mathbb{N}.
\end{equation}
The general Fourier analysis approach is discussed later in Equation~\eqref{eq:depth_amplitude_reconstruction}).
To estimate the integer factor $n$, we can solve the above equation simultaneously for multiple frequencies $\omega$ and leverage Euclidean division~\cite{xia2007phase}. More advanced phase unwrapping algorithms also take into consideration spatial depth information~\cite{lawin2016efficient} and spectral information~\cite{kirmani2013spumic}. We can then convert this phase estimate $\phi$ to depth as $z=\phi c/4\pi\omega_p$, where $c$ is the speed of light.

\begin{figure}[t]
    \centering
    \includegraphics[width=\linewidth]{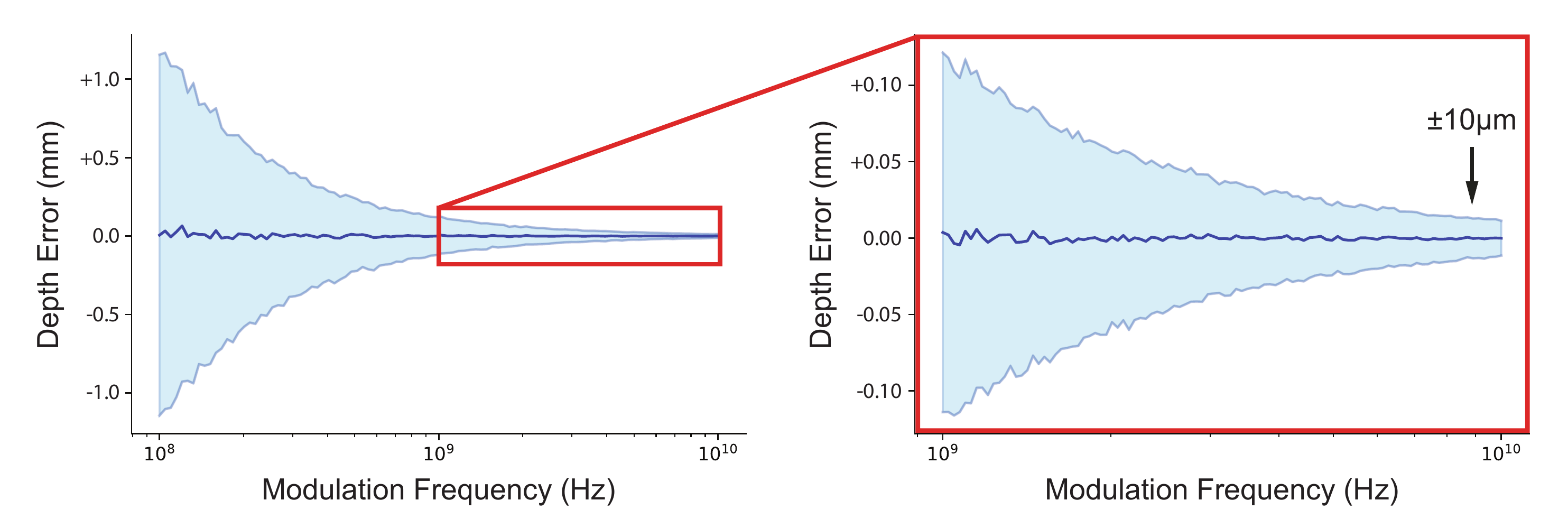}
    \caption{\label{fig:flying_pixels}%
     Illustration of depth estimation error versus modulation frequency. For each frequency we simulate 1000 samples of a point at a depth of exactly 10mm with added Skellam-Gaussian noise, see text, with constant magnitude. We quantize the simulated measurement to 14 bits (mimicking a 14-bit digital-to-analog conversion), reconstruct the estimated depth via Equation ~\ref{eq:depth}, and plot the resultant mean measurement and standard deviation envelope. We see that at a 100MHz frequency we see millimeter precision, as expected, which increases to 10 micrometer precision for a 10GHz modulation frequency.
    }
    \vspace{-1mm}
\end{figure}

\paragraph{Noise Model}
The model up to this point assumes an ideal reflector, no ambient light, and a perfect imaging system. In practice, however, there exist many sources of error in the depth estimation process. Correlation sensing noise can be subdivided into photon shot, read-out, and amplifier noise, which for a typical multi-bucket difference ToF imager can be well approximated by a Skellam-Gaussian model~\cite{hansard2012time,callenberg2017snapshot}. Our all-optical method, however, does not perform bucketing to produce correlation images and so the noise model is instead Poisson-Gaussian, making the measured correlation value $\tilde C_{\psi}(\rho)$
\begin{align}\label{eq:noise_model}
    &\tilde C_{\psi}(\rho) = C_{\psi}(\rho) + \eta_{P} + \eta_{G} \nonumber \\
    &\mathrm{where} \quad \eta_P \sim P(\lambda = C_{\psi}(\rho)), \quad \eta_G \sim \mathcal{N}(\mu, \sigma).
\end{align}
Here $P(\lambda = C_{\psi}(\rho))$ is a sample from the Poisson distribution with rate parameter $\lambda = C_{\psi}(\rho)$, and $\mathcal{N}(\mu, \sigma)$ is a Gaussian distribution with mean $\mu$ and standard deviation $\sigma$.

\begin{figure}[t]
	\centering
	\vspace{-4mm}
	\includegraphics[width=1.03\linewidth]{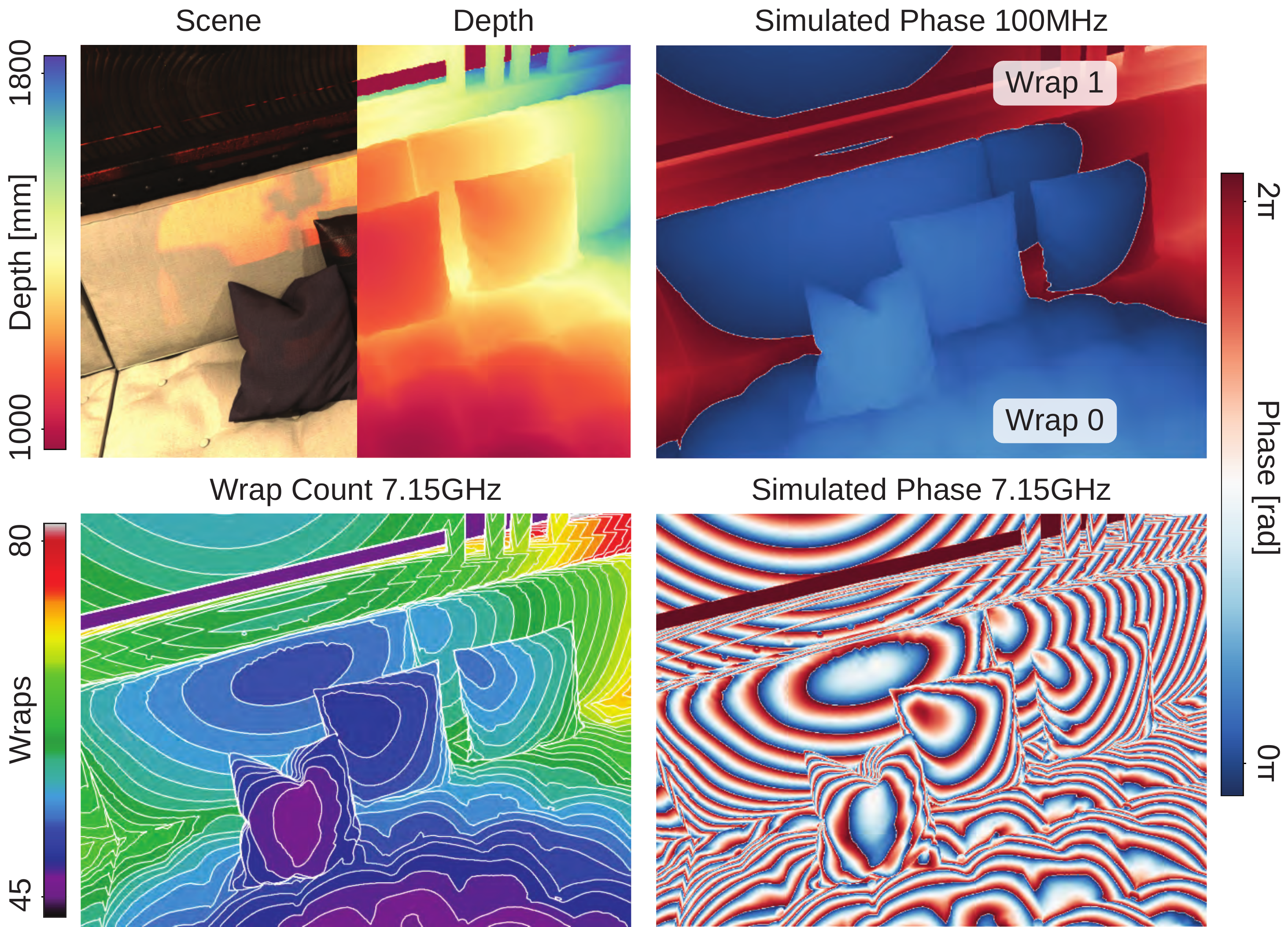}
	\caption{\label{fig:mhz_ghz_wraps}%
    Simulated measurements for a 100MHz ToF system, which exhibits only a single phase wrap, and a 7.15GHz system which experiences 35 wraps.
	}
	\vspace{-4mm}
\end{figure}
\paragraph{Modulation Frequency}
As we noted earlier in Equation~\ref{eq:detected}, the round-trip path of the amplitude-modulated illumination imparts on it a $\phi$ phase shift, which we subsequently use to measure depth $z=\phi c/4\pi\omega_p$. Setting $t=0, \tilde{\beta} = 0$ and $\omega=100MHz$ (a common modulation frequency in conventional ToF cameras) in Equation~\ref{eq:detected}, we observe a 0.0009\% signal difference for a 1mm change in depth $z$.
See Figure~\ref{fig:flying_pixels}.
This means we would practically not be able to discern millimeter scale features on object surfaces for a setup with this modulation frequency. To achieve resolutions well below a mm in precision, we must go to gigahertz frequency, where the above example repeated for $\omega=8GHz$ leads to an easily detectable 5.6\% difference in signal amplitude. For frequencies beyond 10~GHz and 14 bit quantization, the achievable precision reduces to 10 micrometers which we will explore in the remainder of this work. 
\begin{figure*}[h!]
  \centering
  \includegraphics[width=\linewidth]{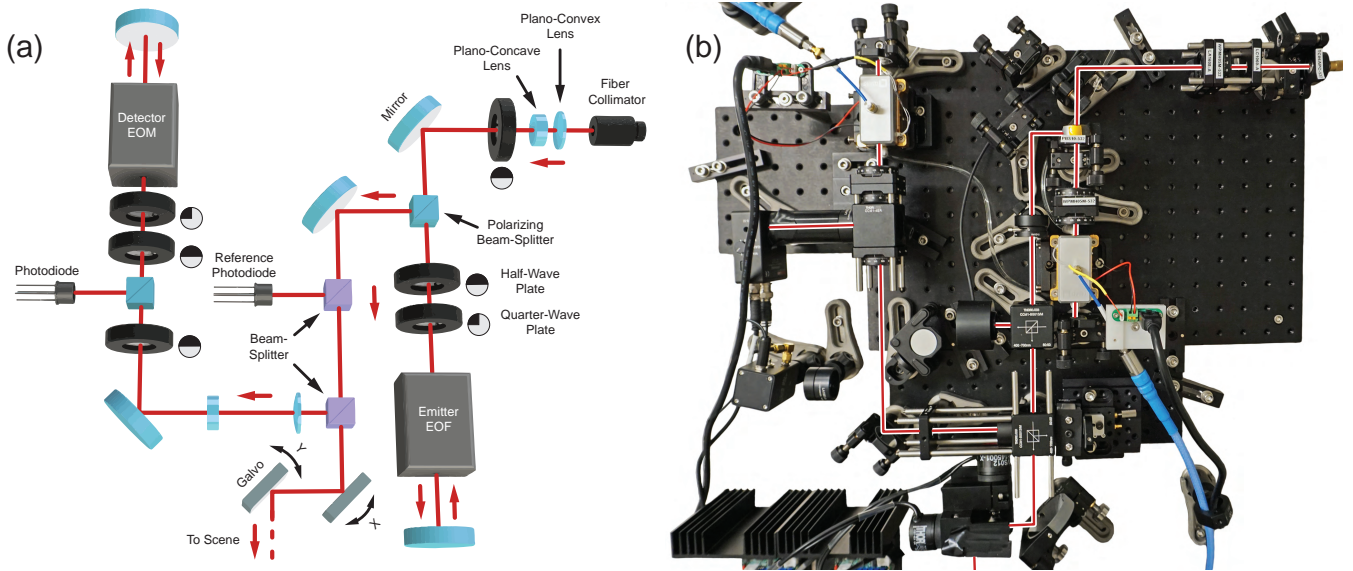}
  \caption{\label{fig:setup_layout}%
  We propose an all-optical free-space AMCW ToF imaging systme using polarizing optics and resonant EOMs. (a) The schematic diagram of our imager realized as in (b) shows the light path from a laser source to a scene, and back to the photodiode. See text for details.
  }
  \vspace{-4mm}
\end{figure*}

\section{Overview of All-Optical Freespace Homodyning}
\label{sec:method}
As described in the previous section, using GHz frequency theoretically enables an order of magnitude improved precision in depth estimation compared to MHz counterparts. However, realizing this depth resolution has been prohibited by \emph{two technical challenges: modulating at GHz rates, and unwrapping the measured phase estimates, see Fig.~\ref{fig:mhz_ghz_wraps}}. We lift these limitations in this work.

First, stable GHz intensity modulation of the illumination and demodulation \emph{before} detection with high modulation contrast is challenging. Unfortunately, analog modulation after photo-conversion or with fiber-coupling suffers from the high noise of ultra-fast photodiodes or large coupling losses. Instead, we propose to perform the correlation computation \emph{optically in free-space} computing correlation before photo-conversion. To this end, we design a novel GHz ToF imaging system that combines electro-optic modulators (EOMs), polarizing optics, an intensity photodiode and analog integration circuits. Figure~\ref{fig:setup_layout} illustrates the proposed imaging system, which \emph{optically computes correlation} relying on polarization modulation in the EOMs. While doubling the frequency of the optical carrier is well known in optics, we propose a novel frequency doubling of the RF intensity modulation.

Second, moving away from MHz modulation rates in existing correlation ToF imagers to GHz rates makes phase unwrapping an obstacle. While MHz correlation ToF imagers have wavelengths on the order of meters, resulting in only a handful of wraps from 0-3, our 7.15GHz modulation frequency corresponds to a wavelength ~4.2cm, resulting about 2*200cm/4.2cm ~ 100 wraps for an operating range of 0-2m, see Fig.~\ref{fig:mhz_ghz_wraps}. At the same time, multi-frequency unwrapping methods, including micro-frequency shifting~\cite{gupta2015phasor}, fail because of the narrow bandwidth of just 20MHz of the resonant electro-optical modulators which renders look-up table based unwrapping (even if inefficient) unusable. We tackle this challenge by introducing a neural network, inspired by modern segmentation networks that routinely can distinguish 100+ image classes. In contrast to semantic segmentation, we exploit that adjacent classes are often highly correlated.

In the next section, we will first describe this computational reconstruction method, before detailing the proposed free-space modulation approach in the subsequent sections.

\section{Neural Phase Unwrapping}
\label{phase_unwrapping}

\begin{figure*}[t]
    \centering
    \includegraphics[width=\linewidth]{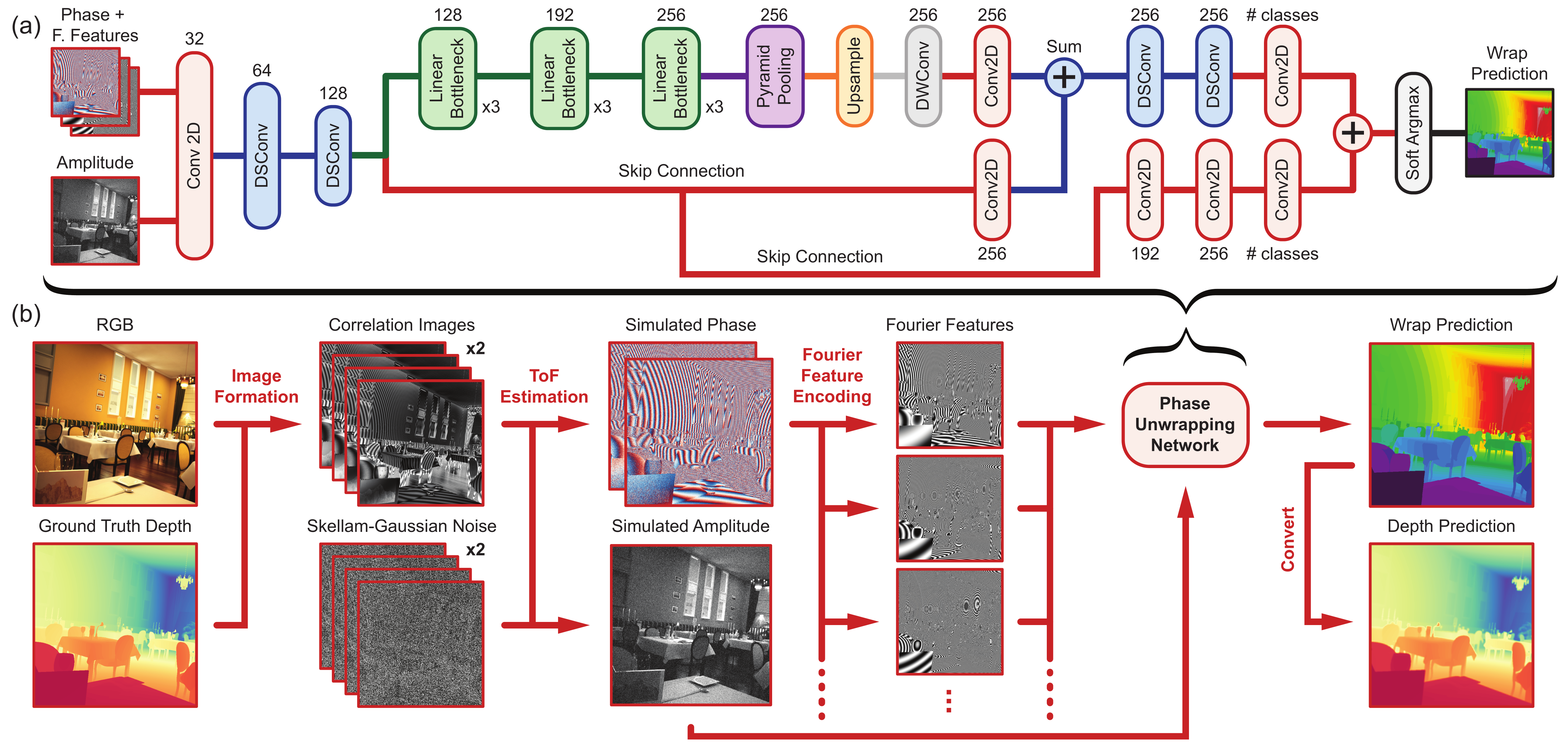}
    \caption{\label{fig:phase_unwrapping_pipeline}%
    Top: The network design used to estimate phase unwrapped measurements. We pose the phase unwrapping problem as a soft ordinal classification problem, and train the network to map the amplitude and phase measurements to wrap counts. We use Fourier feature encoding of the phase measurements to facilitate network learning of high-frequency features and Fourier analysis of the underlying ToF signal.Bottom: Illustration of the phase unwrapping training pipeline. We simulate the amplitude and phase measurements with Poisson-Gaussian noise to ensure a noise-robust phase unwrapping network.
    }
    \vspace{-4mm}
\end{figure*}
Our all-optical ToF imaging system acquires the correlation measurements $C_\psi$ at multiple phase shifts $\psi_{1,...,N}$ between reference signal and demodulation signal.
We then estimate wrapped phase, amplitude, and bias of the correlation signals using a per-pixel Fourier transform detailed below.
However, in order to obtain a final depth map from the obtained phase, we need to unwrap the resulting phase.
As noted in the previous section, the estimation of phase naturally leads a to a $2\pi n, n\in \mathcal{N}$ ambiguity, where we refer to $n$ as the \textit{wrap count} in tens to hundreds. Practically, nearly every pixel in a scene will lie on or close to a phase wrap boundary, as visualized in Figure~\ref{fig:phase_unwrapping_pipeline}. This means even a modicum of noise will alias these measurements into incorrect algebraic solutions for $n$, resulting in severe depth errors.

We tackle this challenge with a novel deep neural network approach that leverages an end-to-end training methodology to achieve robust high-frequency phase unwrapping. In this section, we first describe the phase estimation approach we use to generate the training and test input to this network, followed by a description of the network and training methodology itself.

\subsection{ToF Phase Estimation and Simulation}
\label{sec:phase_simulation}
Given that there do not exist GHz ToF datasets, especially not ones with associated ground truth, we look to simulation to fill our need for training data. Starting with an RGB-D dataset, containing ground truth depth maps $Z\in\mathbb{R}^{H \times W}$ (in mm) and images $I\mathbb{R}^{H \times W}$, we first calculate ground truth phase as $\Phi = (Z 4\pi \omega)/c$, where $\omega$ is modulation frequency and $c$ is the speed of light. Following Equation~\eqref{eq:correlation} we then simulate N ToF correlation images $C_1, ... C_N$ as
\begin{align}\label{eq:correlation_images}
    C_\psi = GI_G(0.5+cos(\Phi + \psi)/\pi)T,
\end{align}
where $G$ is sensor gain, $T$ is integration time, and $I_G$ is the green channel of the image (meant to emulate the green laser in the experimental prototype). To simulate measurement fluctuations, we add Poisson-Gaussian noise $N_{P}=\mathcal{P}^{H \times W}(\lambda=C_\psi) + \mathcal{N}^{H \times W}(\mu, \sigma)$, as outlined in Equation~\eqref{eq:noise_model}, to produce $\tilde C_\psi = C_\psi + N_{SG}$.

We note again that a typical correlation difference imager noise model would be Skellam-Gaussian~\cite{hansard2012time}, however our all-optical design has no photon bucketing and subsequently encounters only Poisson-Gaussian noise. Each correlation image is perturbed individually, and we then recover phase $\tilde \Phi$, amplitude $\tilde A$, and offset $\tilde B$ from these measurements as
\begin{align}\label{eq:depth_amplitude_reconstruction}
     \tilde \Phi = \mathrm{angle}(\mathcal{F}_2 (C_\psi)),
     \tilde A = 2|\mathcal{F}_2 (C_\psi)|,
     \tilde B = |\mathcal{F}_0(C_\psi)|,
\end{align}
where $\mathrm{angle}(\cdot)$ is the phase angle of a complex number and $\mathcal{F}_i(\cdot)$ is the $i$-th complex value of the Fourier transformed signal, therefore $\mathcal{F}_0$ being the DC component.
This process is repeated for every additional modulation frequency $\omega_i$, and the arrays are stacked to form the raw multi-frequency measurements. As a result of the above Fourier recovery, $\tilde \Phi \in [0, 2\pi]$ is phase wrapped and requires further processing before it can be transformed back to depth.

\subsection{CRT Unwrapping}
Many multi-frequency phase unwrapping methods for ToF fundamentally either weigh Euclidean division candidates~\cite{bioucas2009multi, droeschel2010multi, lawin2016efficient} or use CRT-like frequency-space lookup tables~\cite{gupta2015phasor} to estimate wrap counts. We thus follow with a description of CRT-based phase unwrapping as it gives insight into the logic behind, and failure cases of, such methods.

While there exist single-frequency phase unwrapping~\cite{herraez2002fast} approaches, these suffer from problems of reference ambiguity; if there is no data with zero wraps in the measurement, where do you start unwrapping? If there are phase discontinuities, how many wraps exist between them? CRT-based unwrapping addresses this issue by asking for a minimum of two measurements for any point, at coprime modulation frequencies $\omega_1$ and $\omega_2$ (i.e $gcd(\omega_1, \omega_2) = 1$, where $gcd(\cdot)$ is the greatest common divisor). This means if our wrapped measurements for these frequencies are $\hat \phi_1$ and $\hat \phi_2$ we have a system of equations
\begin{align}\label{eq:CRT_eqs}
    & \phi_1 = \hat \phi_1 + 2\pi n_1, \quad n_1 \in  \mathbb{N}, \nonumber \\
    & \phi_1 = k\phi_2 = k(\hat \phi_2 + 2\pi n_2, \quad n_2) \in  \mathbb{N}, \nonumber \\
     \rightarrow (&\hat \phi_1 - k\hat \phi_2) - 2\pi(n_1 - kn_2) = 0.
\end{align}
Where by the Chinese Remainder Theorem~\cite{pei1996chinese}, for which the method is named, we have that this last equation admits a single set of solutions $(n_1, n_2)$, which allow us to fully disambiguate $\phi_1$. Here $k$ is a constant factor which allows us to convert from the space of $\omega_2$ to $\omega_1$. In practice, given finite precision, \eqref{eq:CRT_eqs} is seldom equal to exactly zero and so we instead seek to solve
\begin{equation}
    \label{minimization}
    \mathrm{minimize}_{n_1, n_2} \left((\hat \phi_1 - k\hat \phi_2) - 2\pi(n_1 - kn_2)\right)^2,
\end{equation}
 for some set of candidates $(n_1, n_2)$. Checking every integer combination proves computationally prohibitive, so we limit ourselves to candidates within some range $[\mathrm{min\_wrap}, \mathrm{max\_wrap}]$ defined by the expected min and max depths in our scene. Even this leads to $O(n^2)$ combinations, so we can further reduce our set to frequency-feasible values. That is if $\omega_1 \approx 2\omega_2$ we test candidates $(n_1, floor(n_1/2) \pm 1)$, where we expect that $\hat \phi_2$ should have approximately half the wraps of $\hat \phi_1$; this greatly reduces the complexity to $O(n)$.

\subsection{Phase Unwrapping Network}
The base CRT algorithm, albeit powerful, is very noise sensitive; one can consider a $\hat \phi_1 - \hat \phi_2 > 0$ where a slight perturbation $\epsilon$ leads to $\hat \phi_1 - \hat \phi_2 - \epsilon < 0$, which admits a completely new integer solution $(n_1,n_2)$. Popular approaches, such as the use of kernel density estimation in the Microsoft Kinect systems~\cite{lawin2016efficient}, weigh a neighborhood of wrapping hypotheses to encourage spatial consistency. Such approaches, however, fail to consider high-level image features such as surface continuities and texture, and do not have defined behavior at GHz frequencies where a single kernel neighborhood may contain multiple phase wraps borders.

With the previous ToF simulator providing abundant training data, we look to a deep convolutional neural network to address these considerations. As inputs to this network we take the noisy wrapped phase $\tilde \Phi$ and amplitude $\tilde A$, and output unwrapped phase and depth outputs. Rather than synthesizing these outputs directly, we are inspired by the CRT wrap candidate approach, and \emph{pose the learning problem as a soft ordinal classification.} We output $C$ class weights for each input pixel, each corresponding to a candidate wrap count. Here $C$ is determined by the maximum expected wrap count for the lowest modulation frequency (reducing class count and problem complexity). We then calculate our final estimate \textit{unwrapped} phase $\hat \Phi$ via a differentiable argmax.

\begin{equation}
    \label{eq:differentiable_argmax}
    \hat \Phi = \sum_{a=0}^{C-1} a\left(\frac{\gamma e^{\hat \Phi_a}}{\sum_{b=0}^{C-1}e^{\gamma  \hat \Phi_b}} \right),
\end{equation}
where $\Phi_i$ is the predicted weights for phase class $i$, corresponding to $n=i$ wraps, and $\gamma$ adjusts the \textit{hardness} of the differentiable argmax. As $\gamma\rightarrow\infty$ this function numerically approaches a true argmax, but can lead to unstable training behavior. Depth can be calculated as $\hat Z = \hat \Phi c/4\pi\omega_{min}$. This differentiable argmax allows for back-propagation through the class-based phase estimation, meaning we are able to use both entropy-based classification losses on the output class weights or standard image losses on estimated phase or depth.

Due to the ordinal nature of the phase counts, we opt for a mixed cross-entropy $\mathcal{L}_{CE}$ and $\ell_1$ loss $\mathcal{L}_{L1}$ to punish mistakes differently based on the distance between the true class and predicted classes.
\begin{align}\label{eq:loss_fns}
    & \mathcal{L} = \mathcal{L}_{CE} + w_{SL1}\mathcal{L}_{L1} \nonumber \\
    & \mathcal{L}_{L1} = |\hat Z - \hat Z| \nonumber \\
    & \mathcal{L}_{CE} = -\sum_{i=0}^{C-1} \Phi_i log(\hat \Phi_i),
\end{align}
The cross-entropy loss allows us to train the network as a classifying image segmentation network, and the smooth $\ell_1$-term provides a distance metric for the classes, penalizing the network for guessing wrap counts $\hat n$ far away from the true $n$. By tuning the weight parameter $w_{L1}$, we can control this class distance penalty.

For our architecture, illustrated in Figure~\ref{fig:phase_unwrapping_pipeline}, we modify the Fast SCNN~\cite{poudel2019fast} image segmentation network, which has demonstrated state of the art results in real-time image segmentation. To encourage the network to learn local frequency unwrapping, rather than overfitting to global scene structure, we reduce its receptive field and add a full resolution skip layer directly to the output.

\begin{figure}[t]
	\centering
	\includegraphics[width=\linewidth]{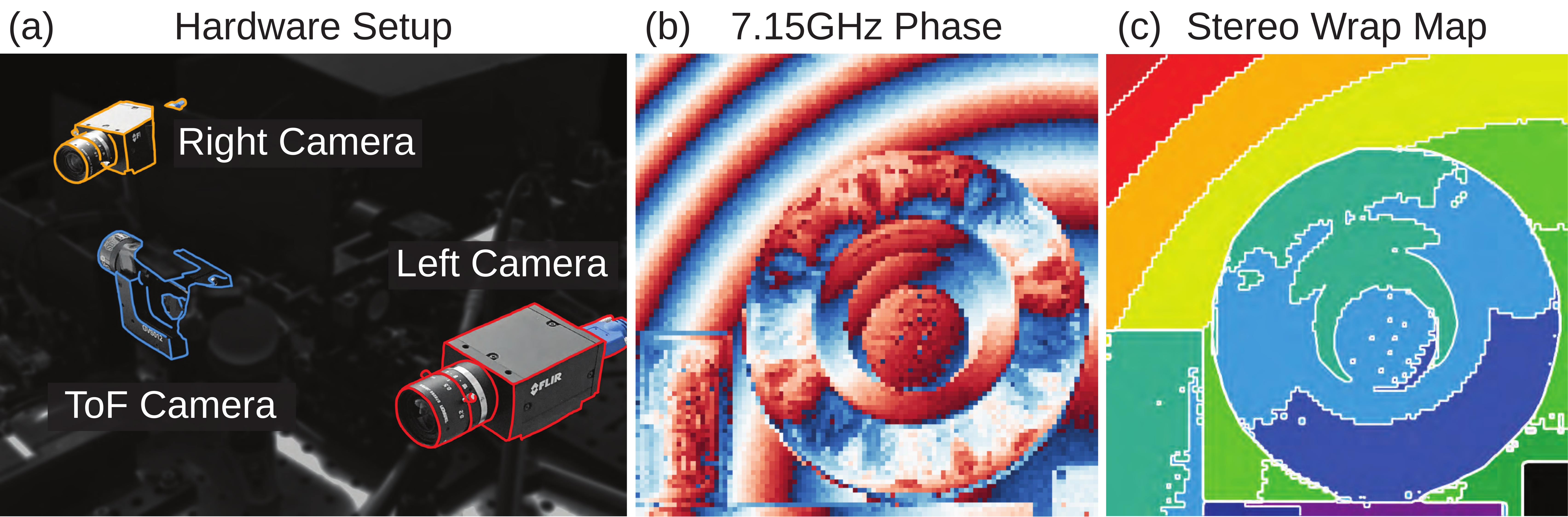}
	\caption{\label{fig:stereo_gt}%
    Active stereo setup to generate ground truth depth data for network finetuning. (a) Dual cameras mounted to the ToF system. (b) Sample 7.15GHz ToF phase map. (c) Corresponding wrap map recovered from stereo depth.
	}
	\vspace{-4mm}
\end{figure}
\subsubsection{Fourier Feature Encoding}
As input to our phase unwrapping network, in addition to measured amplitude $\tilde A$, we use a Fourier feature encoding~\cite{tancik2020fourier} of $\tilde \Phi$
\begin{align}\label{eq:fourier_encoding}
    & \gamma(\tilde \Phi) = [cos(2^0 \tilde \Phi), sin(2^0\tilde \Phi), cos(2^1\tilde \Phi), ..., sin(2^{EC} \tilde \Phi)]^T,
\end{align}
This was used to great success in \cite{mildenhall2020nerf} as a positional encoding method, mapping x,y,z coordinates to a higher dimensional space and improved training for their MLP representation. For our phase unwrapping network the purpose is two-fold. This encoding increases the dimensionality of the input multi-frequency measurements to facilitate learning of high-frequency features, and modulates the correlation values, as we visualize in Figure~\ref{fig:phase_unwrapping_pipeline} (b), allowing for the network to directly learn a Fourier analysis of the underlying ToF signal.

\subsubsection{Finetuning with Active Stereo Supervision}
{
We finetune the proposed network to allow for specialization to the specific noise characteristics of our experimental system and minor deviations of the modulation functions from ideal sinusoids. To this end, as illustrated in Figure~\ref{fig:stereo_gt}, we acquire pseudo ground-truth phase wrap maps by augmenting our system with stereo cameras (other ground-truth acquisition approaches are also possible, we chose stereo for ease of implementation).
Note that this acqusition step is only used for finetuning, and \emph{not employed at inference time}.
We mount two CMOS cameras (FLIR Grasshopper3 GS3-U3-32S4C) to the ToF rig, with 8mm lenses to match our system's FOV.
This immediately creates an auxiliary active stereo system to recover coarse scene depth, without additional captures but during a ToF measurement.
After geometric calibration, we triangulate the position of the ToF laser spot in 3D space with the stereo cameras as we scan the scene.
The estimated depth for each laser spot allows us to generate a ground truth wrap map.
We perform network finetuning on a diverse fine-tuning dataset of captured scenes with stereo measurements, which are all \emph{withheld from the experimental validation} section.
For details on the finetuning, we refer to the Supplemental Document.
} 

\begin{figure}[t]
	\centering
	\includegraphics[width=0.9\linewidth]{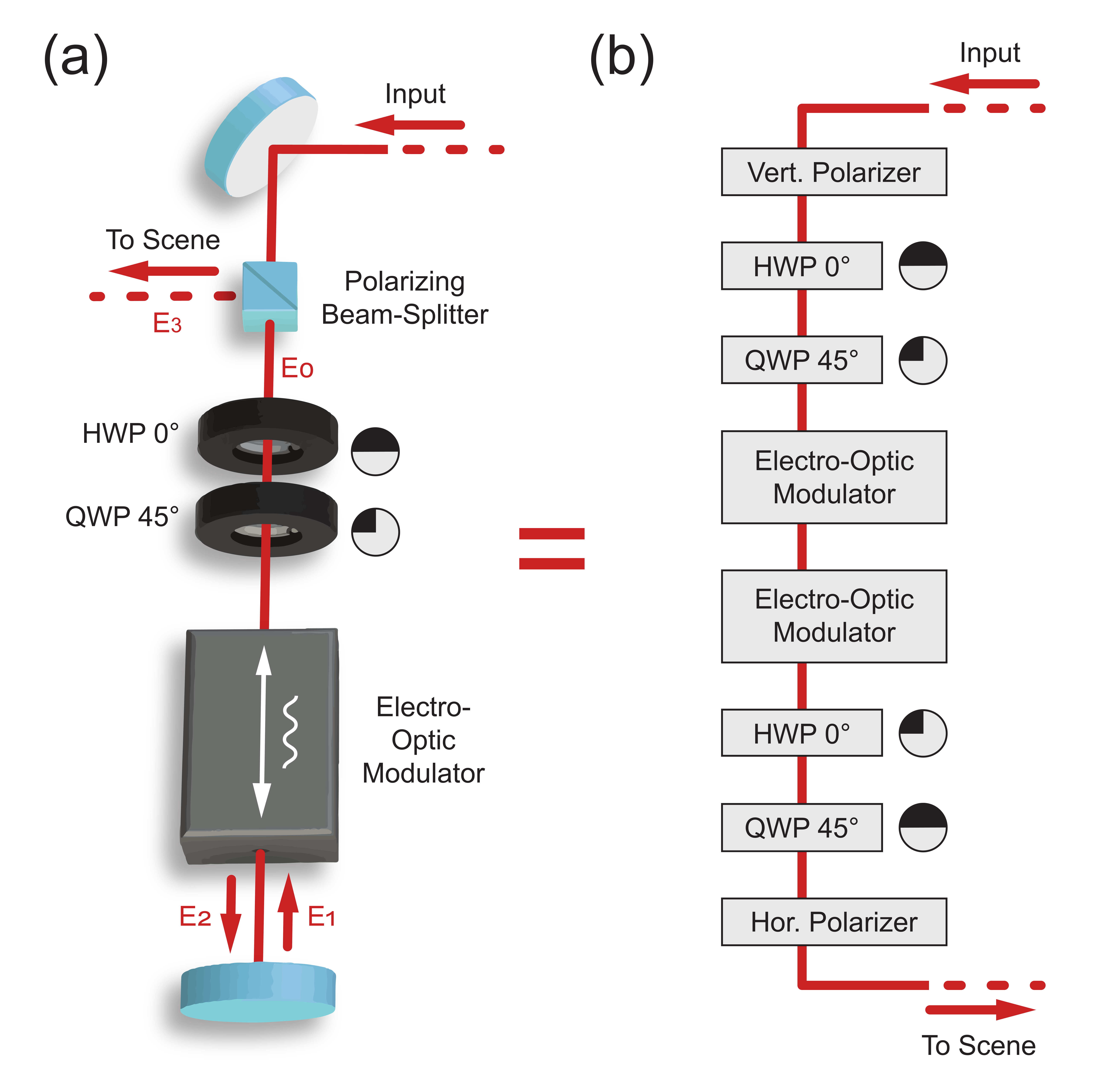}
	\caption{\label{fig:equivalent_design}%
		GHz intensity modulation with polarizing optics and an EOM. (a) We implement the GHz intensity modulation of incident light by using polarizing optics and a EOM. Incident light becomes linearly polarized after a PBS and further polarization modulated by a HWP and a QWP. An EOM with a sinusoidal voltage applied shifts phases of the horizontal and vertical polarization components.
The light then is returned by a mirror distanced at half the modulation wavelength, and returns back to the EOM, the QWP, the HWP, and the PBS.
The combination of forward and reverse paths results in (b) the optical intensity modulation of incident light at GHz frequency with unrolled polarization modulation, see the text for details.
	}
	\vspace{-2mm}
\end{figure}

\section{Computing Correlation Optically with Frequency-Doubling}\label{sec:modulation}

Figure~\ref{fig:setup_layout} illustrates the proposed imaging system. The system emits GHz intensity-modulated light to a scene using an emission module consisting of a laser, polarizing optics, and an EOM. Light is reflected by a scene point and a small portion returns to our detection module with a time delay. In the detection path, another EOM with polarizing optics \emph{optically} computes the correlation between the reflected light and a reference signal without quantization at GHz frequency. The correlated signal is then detected by a photodiode which integrates the photocurrent before analog-to-digital converter.

To this end, we revisit linear electro-optic modulation (EOM) to modulate light intensity at GHz frequencies.
Modern EOMs modulate phase, amplitude, polarization of light with a controllable refractive indices of a bulk crystal by applying electric field to the crystal, perpendicular to the direction of light propagation, according to the electro-optic Pockel's effect~\cite{yariv1967quantum}.
To mathematically model the effect of EOM, we rely on Jones vector and Jones matrix.
The Jones vector is a $2\times1$ vector that describes the amplitude and phase of horizontal and vertical polarization components.
As such, the corresponding Jones matrix describes the change of the polarization state of light with a $2 \times 2$ matrix that can be multiplied to a Jones vector.
We refer the reader to Collet~\shortcite{collett2005field} for a review on Jones calculus.
Specifically, we rely on a Jones matrix for describing the EOM which shifts the horizontal and the vertical polarization waves of light by an amount dependent on the applied voltage $V$ as
\begin{equation}
	B(V) = \begin{bmatrix} e^{-iV/2} & 0  \\ 0 & e^{iV/2} \end{bmatrix},
\end{equation}
where $V$ is the time-varying voltage function
\begin{equation}
\label{eq:voltage}
V=\eta \cos(\omega t - \phi).
\end{equation}
Here $\eta$ is the modulation power, $\omega$ is the voltage modulation frequency, $\phi$ is the modulation phase.

\paragraph{Custom GHz Resonant Modulators}
While EOMs have the advantage of optically modulating light without any quantization, conventional EOMs typically operate at MHz modulation bandwidths.
To reach the GHz modulation band, we have developed custom EOMs with resonant MgO-doped Lithium Niobate crystals, see also~\cite{rueda2016efficient}, that achieves $> 50\%$ modulation contrast for free-space modulation at 7.15~GHz with a few volts.

We note that existing amplitude modulators available on the market were either fiber-coupled or did not achieve GHz modulation envelopes. We had to develop custom crystals that support 7GHz EOM with high modulation contrast at 532nm and non-trivial GHz electronics that perform synchronized phase shifting, both of which are not off-the-shelf. The modulators were fabricated by Qubig GmbH. Our resonant EOM has a housing which forms a resonant cavity that produces standing waves of a modulating frequency causing a change in the refractive index of the crystalline material.

\subsection{GHz Intensity Modulation using Polarization}
Our custom resonant EOM delays the phase of horizontal and vertical components of light at the frequency $\omega$.
We exploit these polarization-dependent phase shifts to implement intensity modulation of incident light.
Specifically, we use the following polarization optics: a polarizing beamsplitter (PBS), a half-wave plate (HWP), a quarter-wave plate (QWP), and a mirror as shown in Figure~\ref{fig:equivalent_design}.
Incident light enters a PBS turning light into vertical linear polarization as
\begin{equation}
	E_{0} = A \begin{bmatrix} 0 \\ 1 \end{bmatrix},
\end{equation}
where $A$ is the amplitude of the incident light.
The polarization state of the light is then modulated by a HWP and a QWP followed by a EOM at a given voltage $V$ as
\begin{equation}
E_{1} = B(V)Q(\theta_q)H(\theta_h)E_{0},
\end{equation}
where the Jones matrices of the HWP and the QWP oriented at angle $\theta_h$ and $\theta_q$ are defined as
\begin{align}\label{eq:wp}
        H(\theta_h) = e^{-i\pi/2}&\left[ {\begin{array}{*{20}{c}}
{{{\cos }^2}\theta_h  - {{\sin }^2}\theta_h }&{2\cos \theta_h \sin \theta_h } \nonumber \\
{2\cos \theta_h \sin \theta_h }&{{{\sin }^2}\theta_h  - {{\cos }^2}\theta_h }
\end{array}} \right],\\
        Q(\theta_q) = e^{-i\pi/4}&\left[ {\begin{array}{*{20}{c}}
{{{\cos }^2}\theta_q  + i{{\sin }^2}\theta_q }&{(1-i)\cos \theta_q \sin \theta_q } \nonumber\\
{(1-i)\cos \theta_q \sin \theta_q }&{{{\sin }^2}\theta_q  +i {{\cos }^2}\theta_q }
\end{array}} \right].
\end{align}
The exitant light from the EOM then propagates in free-space by the distance corresponding to the half of the modulation wavelength $c/\omega$ where a mirror is placed, resulting in the change of Jones vector as
\begin{equation}
E_{2} = M E_{1},
\end{equation}
where $M$ is the Jones matrix of a mirror
\begin{align}\label{eq:mirror}
        M = \left[ {\begin{array}{*{20}{c}}
1 & 0 \\
0 & -1
\end{array}} \right].
\end{align}
Light travels again back to the EOM, the QWP, and the HWP and the PBS picks up the vertical linear polarization component of the light.
Setting the HWP and the QWP angles as $\theta_q=11.25^\circ$  $\theta_q=45^\circ$, we obtain the output light
\begin{align}
E_{3} &= L_h H(-\theta_h)Q(-\theta_q)E_2 \nonumber \\
& = L_h H(-\theta_h)Q(-\theta_q)B(V)MB(V)Q(\theta_q)H(\theta_h)E_0 \nonumber\\
& = \left[ {\begin{array}{*{20}{c}}
\frac{(1+i)e^{-iV}(i+e^{2iV})}{2\sqrt{2}}& \frac{e^{-iV}((1+i)-(1-i)e^{2iV})}{2\sqrt{2}} \\
0 & 0
\end{array}} \right]   E_0 \nonumber \\
& = \left[ {\begin{array}{*{20}{c}}
\frac{i(\cos V + \sin V)}{\sqrt{2}}& \frac{i(\cos V - \sin V)}{\sqrt{2}} \nonumber \\
0 & 0
\end{array}} \right]   E_0 \\
& = A\left[ {\begin{array}{*{20}{c}}
{\frac{{i(\cos V - \sin V)}}{{\sqrt 2 }}}\\
0
\end{array}} \right] ,
\end{align}
where $L_h$ is the Jones matrix of the horizontal linear polarizer
\begin{align}\label{eq:HLP}
        L_h = \left[ {\begin{array}{*{20}{c}}
1 & 0 \\
0 & 0
\end{array}} \right].
\end{align}

We apply the squared magnitude to $E_{3}$, resulting in the modulated intensity $I(V)$ as
\begin{align}\label{eq:out_light}
  I(V) = |E_{3}|^2 = \frac{A^2}{2} (1-\sin 2V).
\end{align}
Equation~\eqref{eq:out_light} indicates that the output intensity of light is a function of the voltage $V$ applied to the EOM.
As we supply a time-varying sinusoidal voltage to the EOM as in Equation~\ref{eq:voltage}, we arrive at the time-varying intensity-modulated light as
\begin{align}\label{eq:out_light_time}
  I(t) &= \frac{A^2}{2} (1-\sin(2\eta\sin(\omega t  - \phi))) \nonumber \\
  &\approx \frac{A^2 }{2} (1-2\eta\sin(\omega t - \phi)).
\end{align}
The last approximation is based on the Taylor expansion assuming the small modulation power $\eta$.
The applied voltage to the EOM has GHz modulation frequency $\omega$, enabling effective all-optical GHz modulation of light intensity. We refer to the Supplemental Material for additional detail.

Equation~\eqref{eq:out_light_time} describes the high-frequency intensity modulation realized by our free-space optical setup shown in Figure~\ref{fig:equivalent_design}.
This optical configuration serves as a \emph{building block for both illumination and detection modules} in our imaging system.
In the illumination module, we input continuous laser light into the EOM, resulting in sinusoidally intensity-modulated light emitted into the scene as $p$ in Equation~\eqref{eq:light_source}.
For the detection module, the returned amplitude-modulated light $\tilde{p}$ from the scene is demodulated by an additional intensity modulation with the reference signal $r$, recall Equation~\eqref{eq:correlation_homodyne}, and we \emph{optically} multiply $r$ and $\tilde{p}$ before integration on the detector.

\paragraph{Double-Frequency Modulation}
Even though the voltage modulation frequency $\omega$ is limited to a narrow modulation band in our resonant EOM, we can modulate at the double frequency of $2\omega$ by adjusting the angle of the HWP, $\theta_h$, in front of the EOM. While doubling the frequency of the optical carrier is well known in optics, we note that the proposed frequency doubling of the RF intensity modulation is novel.
In the original operating mode, we set $\theta_h$ as $11.25^\circ$ resulting in the intensity modulation at $\omega$.
For frequency doubling, we rotate the HWP to $\theta_h = 22.5^\circ$. To derive the modulation behavior, we rely on the same Jones calculus from above.
Specifically, changing $\theta_h$ in the polarization in the system of Jones matrices results in the output light $E_3$ as
\begin{align}
E_{3} &= L_h H(-\theta_h)Q(-\theta_q)E_2 \nonumber \\
& = L_h H(-\theta_h)Q(-\theta_q)B(V)MB(V)Q(\theta_q)H(\theta_h)E_0 \nonumber\\
& = \left[ {\begin{array}{*{20}{c}}
\frac{ie^{-iV}(i+e^{2iV})}{2}& -\frac{e^{-iV}(-1+e^{2iV})}{2} \\
0 & 0
\end{array}} \right]   E_0 \nonumber \\
& = \left[ {\begin{array}{*{20}{c}}
i\cos V & -i \sin V \nonumber \\
0 & 0
\end{array}} \right]   E_0 \\
& = A\left[ {\begin{array}{*{20}{c}}
-i \sin V\\
0
\end{array}} \right].
\end{align}
The intensity $I(V)$ is the magnitude square of $E_3$ as
\begin{equation}
\label{eq:out_light_double}
I(t) = |E_3|^2 = \frac{A^2}{2} (1-\cos(2V)).
\end{equation}
Note that the difference of Equation~\eqref{eq:out_light_double} with Equation~\eqref{eq:out_light} is that we have $\cos()$ instead of $\sin()$.
This single difference enables us to arrive at the intensity modulation at double frequency.
After applying the time-varying voltage modulation of Equation~\eqref{eq:voltage}, the time-varying intensity of the output light is
\begin{align}\label{eq:out_light_double_time}
  I(t) &= \frac{A^2}{2} (1-\cos(2\eta\sin(\omega t  - \phi))) \nonumber \\
  &\approx \frac{A^2 }{2} \eta^2 \sin^2(\omega t - \phi) \nonumber \\
  & = \frac{A^2 }{4} \eta^2 (1 - \cos(2\omega t - 2\phi))
\end{align}
Note that we use the same Taylor expansion with small modulation power $\eta$ in the second approximation.
Equation~\eqref{eq:out_light_double_time} shows that we can obtain doubled frequency modulation of $2\omega$ with reduced amplitude by the factor of four compared to the single-frequency mode at $\omega$ -- only by changing the polarization optics instead of the electro-optical modulation itself.
\begin{figure}[t]
	\centering
	\vspace{-4mm}
	\includegraphics[width=\linewidth]{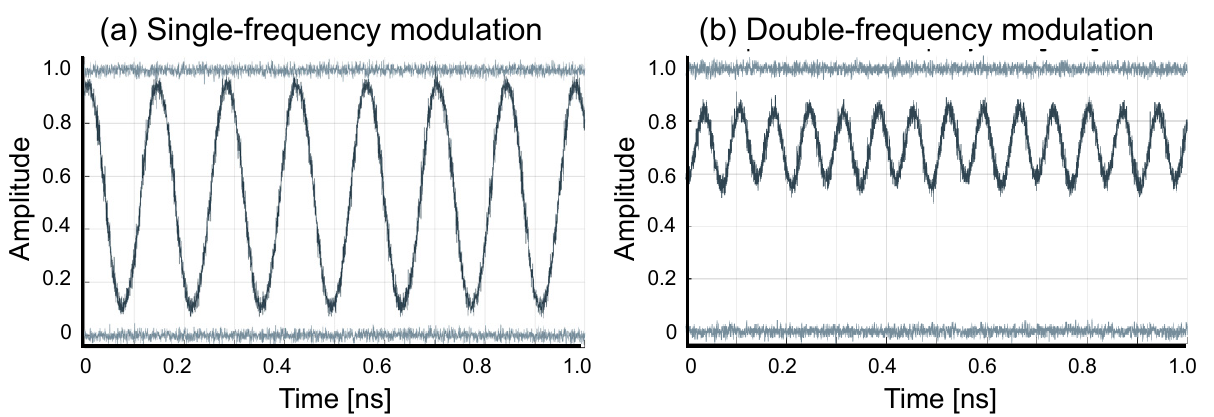}
	\caption{\label{fig:validation_eom}%
		We validate our GHz intensity modulation module by capturing the reflected GHz intensity from a scene mirror at a fixed position.
        (b) When the HWP $\lambda/2$ is oriented at $11.25^\circ$, we achieve the single intensity modulation frequency of $\omega=7.15$GHz.
        (c) Changing the HWP angle to $22.5^\circ$ enables frequency doubling at $2\omega=14.3$GHz.
	}
	\vspace{-4mm}
\end{figure}

\paragraph{Validation of GHz Intensity Modulation}
We validate our GHz intensity modulation module consisting of a PBS, a HWP, a QWP, a EOM, and a mirror.
We emit laser light from our illumination module to a scene mirror at a fixed position and \emph{directly} capture the intensity of the modulated light steered onto a GHz photodetector, see Supplemental Material for the measurement configuration. Figure~\ref{fig:validation_eom} demonstrates the effective GHz intensity modulation with high modulation contrast at two different HWP angles of $11.25^\circ$ and $22.5^\circ$ corresponding to the modulation frequencies of $\omega=7.15$\,GHz and $2\omega=14.3$GHz.

\subsection{Coaxial Spatial Imaging}
Equipped with the intensity modulation block, we design a coaxial imaging system with an illumination and a detection module, see Figure~\ref{fig:setup_layout}.
For the illumination module, we opt for coherent laser illumination at 532~nm (for eye-safe lab operation of the prototype) followed by a GHz intensity modulation block. A further GHz modulation block is used for the detection module combined with an avalanche photodiode (APD) that is used for intensity sensing.
Using a non-polarizing beamsplitter, we share the same path for the output light to a scene and the detected light from a scene, improving the signal-to-ratio of the system.
For 2D spatial scanning, we use a 2-axis galvonometer in front of the beamsplitter, shown in the bottom of Figure~\ref{fig:setup_layout}. Although the proposed free-space modulation method is not limited to co-axial scanning, the beam-steered acquisition effectively eliminates most multi-path interference, which we neglect in the remainder of this work.

\paragraph{Analog Signal Integration}
We use a conventional avalanche photodiode with a gain $G$ to detect the correlation signals from the detection module without any quantization involved.
This generates analog photocurrent which is then low-pass filtered with an electrical filter and a resistor-capacitor (RC) circuit that further integrates the constant correlation input signal over an exposure time $T$.
We read out the analog signal with an ADC with 14bit quantization.
This results in the digital read of
\begin{align}\label{eq:measured_correlation}
    C_\psi = Q\left(G\int_{\rho}^{\rho+T} {\tilde p( {t - \tau } )r( {t} ) \, \mathrm{d}t} \right)
     = Q\left(\frac{\tilde{\alpha}}{2}\cos(\psi - \phi) + TK \right),
\end{align}
where $Q$ is the 14bit ADC quantization operator, $\phi$ is the phase of the emitted light as in Equation~\eqref{eq:light_source}, and $\psi$ is the phase of the reference signal $r$ shown in Equation~\eqref{eq:correlation_homodyne}.

\section{Experimental Setup}\label{sec:experimental}
For experimental validation, we implement the prototype system illustrated and shown in Figure~\ref{fig:tof_imaging}.
{
While our experimental prototype is currently assembled on an optical breadboard, we note that the EOMs and optics can be integrated in small form factors similar to existing lidar sensors.
}

\paragraph{Illumination Module}
We use a single transverse mode continuous wave laser at 532 nm wavelength (Laser Quantum Gem 532).
The laser beam is coupled with a custom-design single mode high power optical fiber (OZ Optics QPMJ-A3AHPCA3AHPC-488-3.5/125-3AS-1-1) which removes the higher order modal light and produces a uniform Gaussian beam at the output of the fiber, maintaining the $20-30\%$ laser output power.
The light then enters an 2.5$\times$ inverse beam expander (Thorlabs LC1060-A and LA1608-A ) that reduces the beam diameter down to ~0.5\,mm, from ~1.25\,mm to be matched with the desired beam size to our EOM.
The reduced light becomes horizontally linearly polarized by passing through a first PBS (Thorlabs PBS101).
Then, a pair of HWP (Thorlabs WPMH05M-532) and QWP (Thorlabs WPQ05M-532) modulates the polarization state of the beam.
The polarization-modulated light passes through the EOM that operates at the modulation frequency $\omega$. As discussed in Sec.~\ref{sec:modulation} the modulator is not an off-the-shelf product but required to develop a custom resonant crystal which was fabricated by Qubig GmbH.
The light is reflected by a mirror (Thorlabs PF10-03-P01), returning back to the EOM, the QWP, the HWP, and the PBS.
This procedure results in the GHz intensity modulation of light.

The light then passes through a mirror (Thorlabs PF10-03-P01) and a NBS (Thorlabs CCM1-BS013) dividing the incident beam into two beams of equal intensity.
One beam is directed to an integrating sphere (Thorlabs S140C) which measures the intensity of emitted light for a calibration purpose and the other beam passes through another NBS (Thorlabs CCM1-BS013). The purpose of this module is to calibrate intensity fluctuations from the Excel laser by normalizing the signal incident on the detection module. The optical intensity modulation has higher frequency than the integration time of a few milliseconds, which allows compensation after the modulation without error.
It splits the beam again into two paths with equal intensity where one half of the beam is used as the reference beam for interferometric measurement mode (used for precision comparison see Fig.~\ref{fig:experimental_precision}) with a mirror; otherwise beam dumped (Thorlabs LBM1) in a normal intensity-measurement mode.
The other half of the beam is sent to a scene through a mirror (Thorlabs PF30-03-P01) and a 2-axis galvo mirror system (Thorlabs GVS012) for spatial scanning. The emitted CW laser power is $3$\,mW in our system. For photon-efficiency estimates, see Supplemental Document.

\paragraph{Detection Module}
The intensity-modulated light returns from a scene and passes through the galvo mirror system and the mirror followed by a NBS which redirects the beam to the detection module.
We use an $~1.6\times$ inverse beam expander (Thorlabs LA1213-A and Thorlabs LC1060-A) and a mirror (Thorlabs PF10-03-P01) resulting in a beam diameter of ~0.5\,mm and collimated beam accurately entering the detection EOM.
Symmetric to the emission module, we mount a PBS, a HWP, a QWP, an EOM, and a mirror that constitute an optical demodulation of returned light from a scene.
The intensity demodulated light is then captured by an avalanche photodiode (Thorlabs APD440A) with a focusing lens (Thorlabs LA1951-A).
We use a 10~kHz lowpass filter (Thorlabs EF120) resistor capacitor (RC) low pass integrator circuit with RC time constant $t_{RC} = 100 ms$ to integrate the detected photocurrent signal, then passed into an analog-digital-converter (LabJack T7) to sample the signal at up to 24K samples per second. We integrate over 20 samples for a single phase measurement and sample 16 phases corresponding to 13ms integration time a single galvo measurement point.

\paragraph{RF Driver}
To operate the EOMs with a sinusoidal voltage input, we use two custom RF drivers with a high-frequency DDS which are synchronized with an external clock source of a function generator (Siglent SDG2042X). The external clock enables accurate controls the phase of the modulation signal $\phi$. Our driver contains two RF modulators to output an RF signal provided to the EOMs. The RF driver performs frequency locking of an RF output signal to significantly increase the output power and reduce the frequency drifting in the EOM. For further details, refer to the Supplemental Document.

\begin{figure*}[t]
	\centering
	\includegraphics[width=\linewidth]{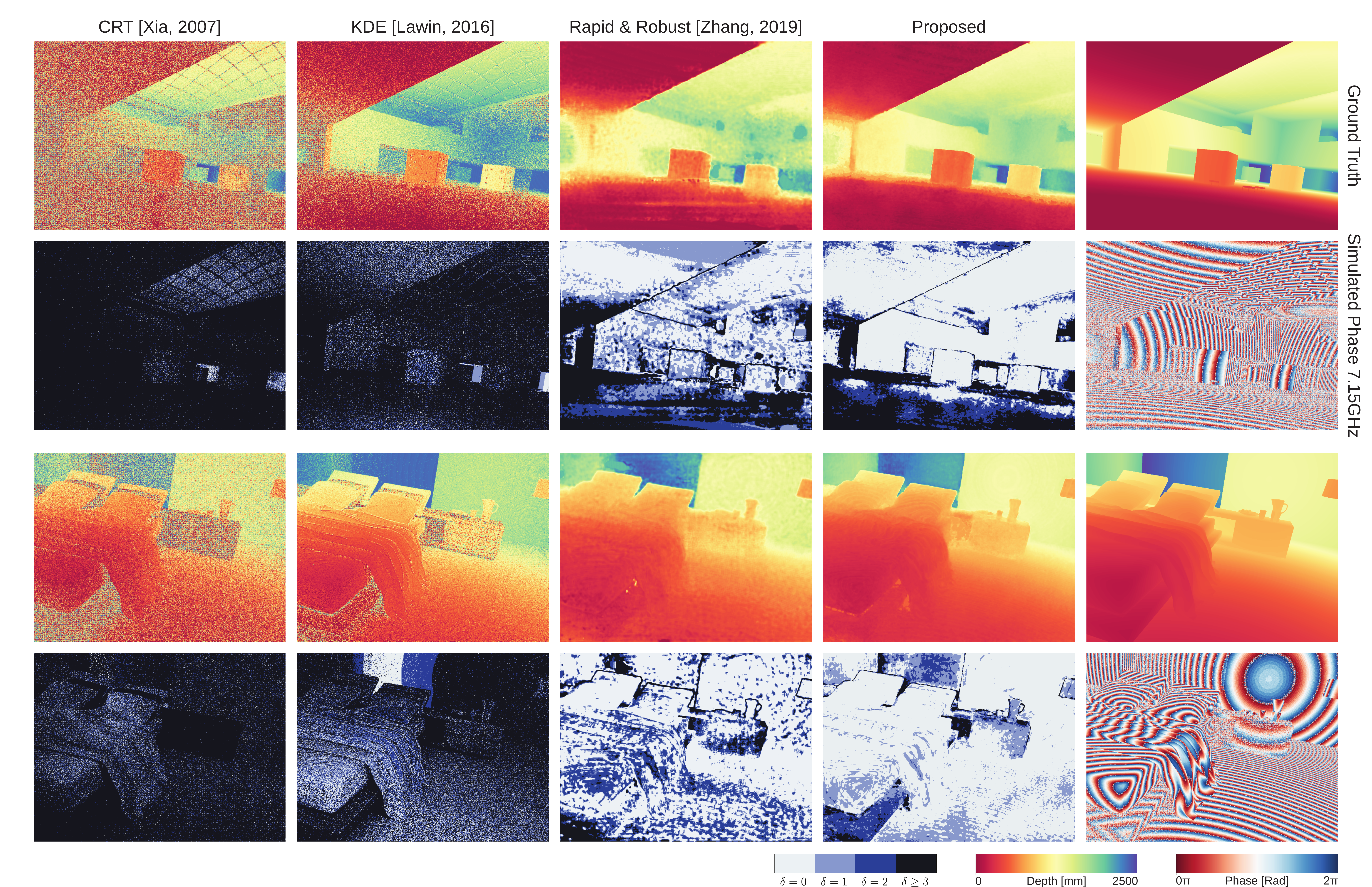}
	\vspace*{-2em}
	\caption{\label{fig:synthetic_results}%
	Phase unwrapping results for comparison to existing conventional and learned methods and our proposed approach. {Analytic solutions of CRT~\cite{xia2007phase} and KDE~\cite{zhang2019rapid} suffer from rapid phase wraps and phase noise. The state-of-the-art neural network method partly overcomes such problems at the cost of smoothed geometry and low-frequency depth artifacts. Our method outperforms the previous methods by recovering both accurate scale and geometric details.} Error map below results corresponds to a visual representation of the $\delta$ metric, see text, in Tables~\ref{table:quantitative_comparison_simulation} and \ref{tab:ablation_config}.
	}
  \end{figure*}
\paragraph{Comparison to RF Demodulation in the Analog Domain}
For comparison of the proposed system with demodulation of a signal after photo-conversion, we add a highspeed GaAs 12GHz photodetector (EOT GaAs PIN Detector ET-4000)  connected to an RF demodulation circuit. This measurement setup can be enabled by flipping a flip-mirror in the optical path, redirecting the scene illumination to the fast photodiode instead of the proposed detection module. The photodiode offered the highest photon-detection efficiency and high-frequency response available to us. The captured photocurrent from the detector is sent as input to an I/Q demodulator consisting of analog microwave electronics as follows. The photodector signal is first amplified and band-pass filtered. Then it enters an RF mixer to be demodulated with the local oscillator (LO) signal from the RF driver. This produces a signal with the difference of the two frequencies and a signal with the sum of the two frequencies. These signals are passed through a low pass filter which removes the higher frequency signal. Then the remaining  homodyne DC signal, is output as two signals, an in-phase component called $I$, and a quadrature phase component $Q$ shifted by 90 degrees. For a detailed circuit design, please see the Supplementary Document.

\section{Assessment}
\label{sec:results}

In this section, we validate our proposed computational time-of-flight method in simulation and with the proposed prototype system. Specifically, we first perform quantitative evaluation of our neural unwrapping approach on a synthetic dataset, following the simulation outline given in Sec.~\ref{sec:phase_simulation}, and compare with other baseline phase unwrapping methods. We then experimentally validate the proposed system quantitatively and qualitative on unseen real-world measurements captured by our experimental prototype.

\subsection{Simulated Analysis }
We fill the lack of existing GHz ToF datasets by using the RGB-D ground truth data from the Hypersim dataset~\cite{roberts:2020} to generate simulated measurements. This dataset contains 77,400 sample image-depth pairs from a total of 461 computer generated 3D indoor scenes. These closely resemble the potential environment in which future GHz ToF systems could be deployed.

\paragraph{Ablation Study}
We conduct an ablation study to validate our choice of Fourier feature encoding and combined loss function. The different ablation configurations and corresponding quantitative results are shown in Table~\ref{tab:ablation_config}, and we refer to Supplemental Document for qualitative results.

\begin{table}[t]
\vspace{-1pt}
\small
\centering
\setlength{\tabcolsep}{2pt}
\begin{tabular}{lccccccccc}
\toprule
			& \multicolumn{3}{c}{{Input}} 	& \multicolumn{2}{c}{{Loss}}&\multicolumn{4}{c}{{Performance (\%)}}\\ \midrule
            & $\tilde \Phi$ 				& $ \gamma(\tilde \Phi)$ 	& $\tilde A$ & $ \mathcal{L}_{CE} $ & $\mathcal{L}_{L1}$ & $\uparrow\delta = 0$ & $\uparrow\delta \leq 1$ & $\uparrow\delta \leq 2 $ & $\downarrow\delta \geq 3 $\\ \midrule \midrule
Proposed					&\checkmark&\checkmark&\checkmark&\checkmark&\checkmark&47.7\%&\textbf{70.3\%}&\textbf{80.0\%} &  \textbf{20.0\%}\\
$ \mathcal{L}_{CE} $ Only   &\checkmark&\checkmark&\checkmark&\checkmark&    -     &\textbf{48.8\%}&66.8\%&75.2\% &24.8\% \\
F. Features           	&\checkmark&\checkmark&    -     &\checkmark&\checkmark&40.2\%&59.6\%&68.8\% &31.2\%\\
Phase Only                  &\checkmark&    -     &    -     &\checkmark&\checkmark&30.3\%&52.3\%&65.7\% &34.3\% \\
			\bottomrule
\end{tabular}
\caption{\label{tab:ablation_config}%
Ablation study configurations and corresponding quantitative results. Here the $\delta$ metric represents the percent of pixels whose prediction is $\delta$ wraps from ground truth wrap count. Up arrow denotes "higher is better", down arrow means "lower is better".}
\vspace{-8mm}
\end{table}

\begin{table}[t]
	\resizebox{\columnwidth}{!}{
	\begin{tabular}{ c c  c  c  c  c}
		\toprule
		Method & $\uparrow\delta=0$ & $\uparrow\delta \leq 1$ & $\uparrow\delta\leq 2$ & $\downarrow\delta \geq 3$ & $\downarrow\delta \geq 10$\\
			\midrule
			\midrule
		Phasor [2015]    		&0.74\%&1.66\%&3.50\%&96.5\%&84.4\%\\
		CRT [2007]   			&9.29\%&14.7\%&19.7\%&80.3\%&56.0\%\\
		KDE [2016]   			&9.46\%&18.56\%&27.0\%&73.0\%&8.93\%\\
		One-Step [2019]			&19.9\%&37.6\%&52.2\%&47.8\%&14.6\%\\
		U-Net [2015]			&21.8\%&45.6\%&64.4\%&35.6\%&10.0\%\\
		Rapid. [2019]			&23.1\%&45.4\%&61.1\%&38.9&9.74\%\\
		Proposed  				&\textbf{47.7\%}&\textbf{70.3\%}&\textbf{80.0\%}&\textbf{20.0\%}&\textbf{3.91}\%\\
			\bottomrule
	\end{tabular}
	}
	\caption{\label{table:quantitative_comparison_simulation}%
	Quantitative comparison table for proposed neural phase unwrapping method and baselines, as evaluated on the synthetic test scenes. $\delta\geq10$ metric added to better quantify outlier performance.
	}
	\vspace{-8mm}
\end{table}

We observe that Fourier encoding leads to a 10 percentage point boost in correct wrap predictions, supporting the theory that the doubly modulated phases provide valuable features during training, possibly in the form of a learned frequency analysis of the underlying measurements. Concatenating the amplitude measurement to the input aids the network with correctly reconstructing fine structures by providing it with addition semantic information, which leads to a bump in performance. Concerning loss functions, we find the model trained on cross-entropy loss alone demonstrates competitive results, validating the choice to represent phase unwrapping as a classification problem. However, when we make use of the differentiable argmax function to directly introduce $\ell_1$ loss on predicted depth, we see a reduction in outliers and an overall smoother final prediction. This reinforces the problem as ordinal classification, where the ordering of classes --- in this case wrap counts --- is significant.

\paragraph{Simulated Results}
We validate our proposed neural unwrapping approach on a synthetic test set and discuss the qualitative and quantitative results. As a baseline, we compare our work against traditional unwrapping methods including the approach used in phasor imaging~\cite{gupta2015phasor}, the algebraic CRT solution~\cite{pei1996chinese, xia2007phase}, and the kernel density method (KDE)~\cite{lawin2016efficient}, which is also used in the Kinect V2 software. We additionally compare to an unmodified U-Net~\cite{ronneberger2015u} baseline and two recent single-step deep learning approaches~\cite{wang2019one, zhang2019rapid}. We note that the U-Net architecture is also used in Su et al.~\cite{su2018deep}.

\begin{figure}[t]
  \centering
  \includegraphics[width=1.015\linewidth]{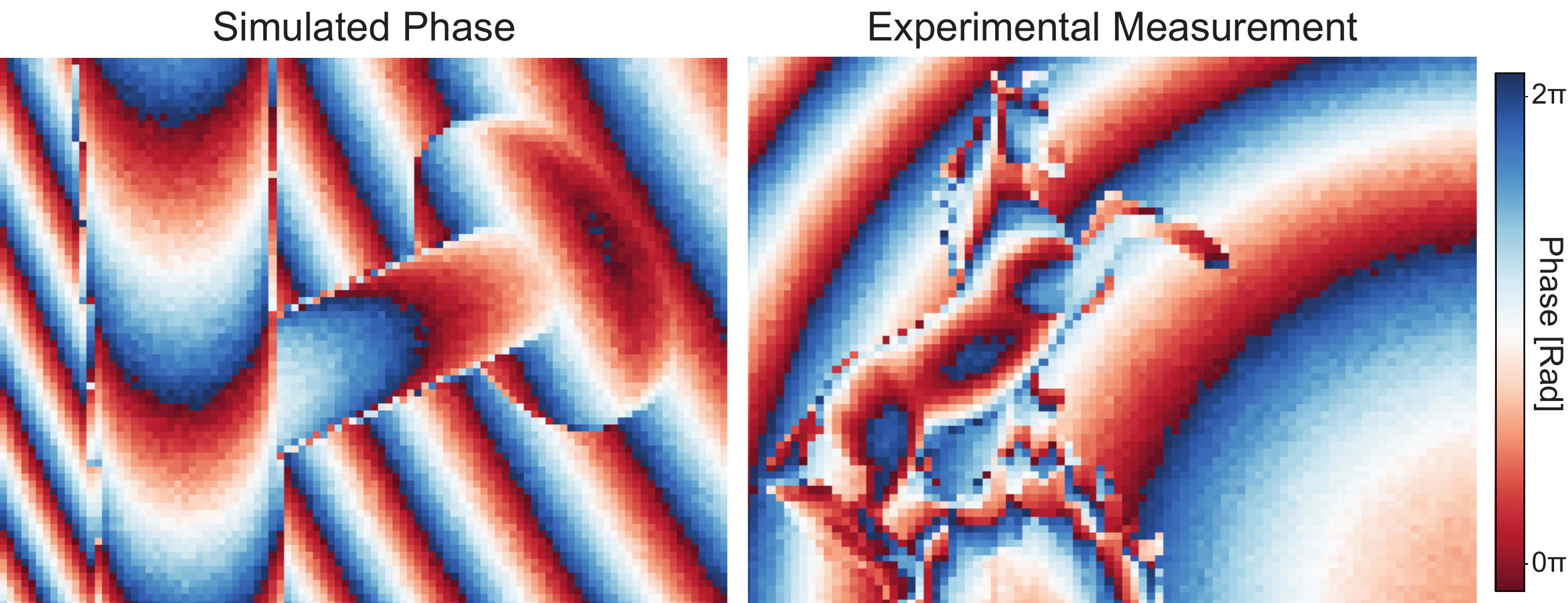}
  \caption{\label{fig:noise_model}%
  Noise matching example. Left: synthetic data with noise parameters $\mu=0, \sigma=1200$ as outlined in \ref{sec:phase_simulation}. Right: experimental measurement.
  }
	\vspace{-2mm}
\end{figure}

For all methods except phasor imaging we simulate measurements for two modulation frequencies, a fundamental 7.15GHz signal and a shifted plus frequency doubled (7.15GHz + 10MHz) $\times$ 2 $=$ 14.32GHz signal. We note that these frequencies correspond to the frequencies we can implement in the experimental setup. For the phasor imaging method, we input 7.15GHz and 7.16GHz simulated measurements, as these are the locally optimal feasible shifts achieved by the optical amplitude modulation system.
We simulate measurements with sensor gain $G=20$ and integration time $T=1000$ms, with noise parameters $\mu=0, \sigma=1200$ chosen to match the signal distribution observed in experimental data as shown in Fig.~\ref{fig:noise_model}.
The models are trained for 1000 epochs each, with 500 samples drawn per epoch, each consisting of a 512$\times$512 image and ground truth depth patch (sampled randomly from the full RGB-D datum). We use an exponentially decaying learning rate schedule with a ratio of $0.995$ per epoch and an initial rate of $1e-3$; training on 3 Nvidia V100 GPUs with a batch size of 12 takes approximately 24 hours. The synthetic test set consists of the 42nd frame of each simulated scene, withheld from both the training and validation sets. We balance the losses by setting $w_{L1}=0.1$, which leads to noticable improvements in smoothness without the classifier's early training behavior. During inference, running on one Nvidia V100 GPU, we achieve an average runtime of 16.5ms$\approx$60FPS per image of size 256$\times$256, and 50ms$\approx$20FPS with the full synthetic image size of 768x1024.
We showcase qualitative performance of our proposed neural unwrapping method and baseline methods including CRT~\cite{xia2007phase}, KDE~\cite{lawin2016efficient} and the next best network-based method~\cite{zhang2019rapid} in Figure ~\ref{fig:synthetic_results}, and refer to the Supplemental Document for additional qualitative comparisons. Table \ref{table:quantitative_comparison_simulation} presents quantitative classification results for the full range of methods in increasingly widening error bands, as well as outlier percentages.

\begin{figure*}[htp]
  \centering
  \includegraphics[width=0.98\linewidth]{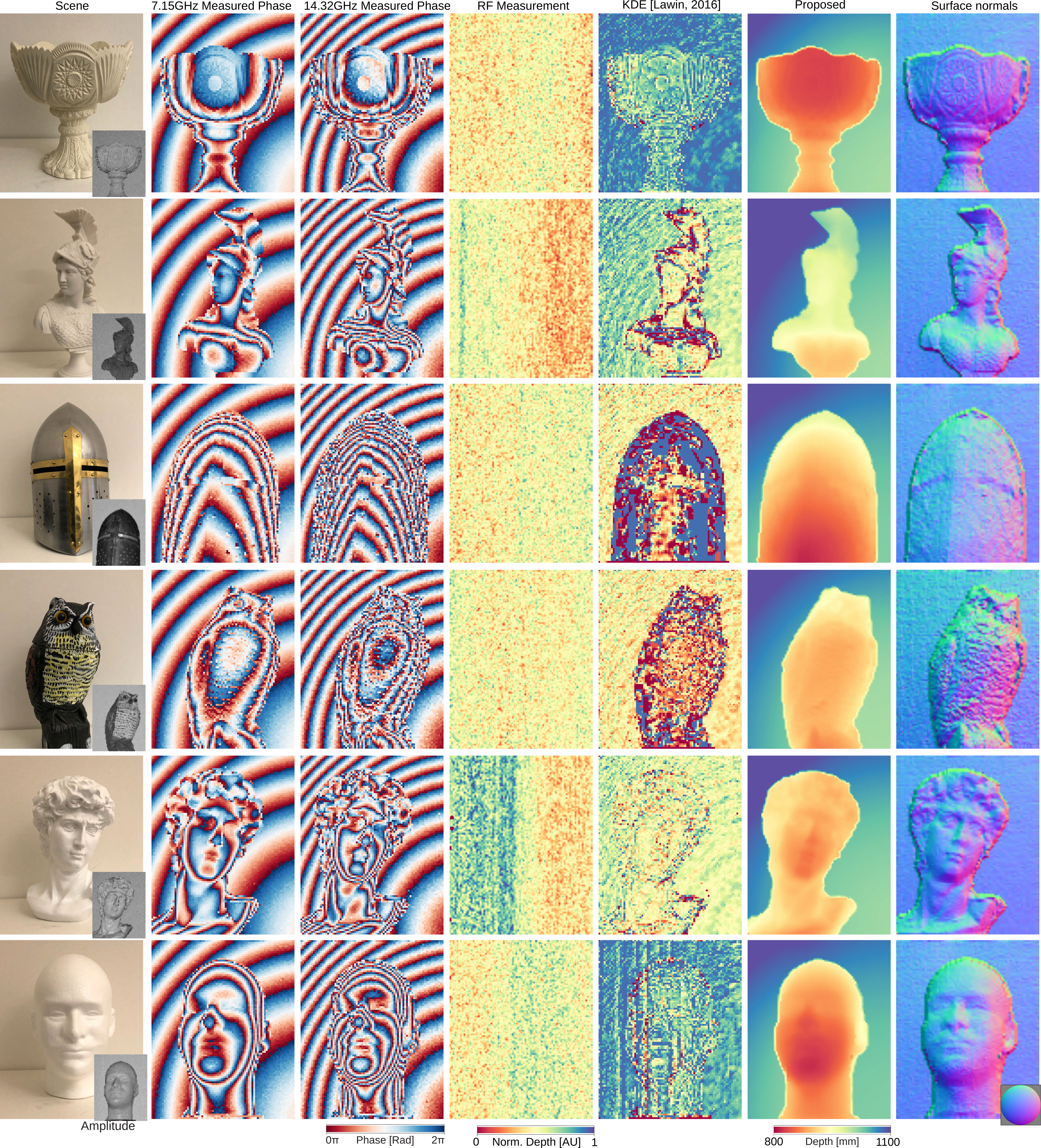}
  \caption{\label{fig:experimental_results}%
  We experimentally acquire macroscopic scenes with our computational ToF system at 7.15 and 14.32 GHz (frequency-doubled 7.16 GHz) frequencies.
  The first column shows an RGB scene photo and the recovered amplitude. While GHz RF demodulation after photo-conversion (see text) struggles with low photon flux of the returned signal, our all-optical correlation imager measures accurate phase maps for both GHz frequencies which encode accurate scene depth information. Obtaining accurate depth from the phase maps, however, is still challenging for existing unwrapping methods due to the high wrapping counts of GHz frequencies. Our neural unwrapping method successfully resolves this issue and enables accurate depth reconstruction.
  }
\end{figure*}

Visually CRT and KDE achieve similar results, as they have similar underlying mechanics for wrap calculation, however KDE's spatial aggregation allows it better tackle noise and make significantly more correct estimates. This is quantitatively confirmed by the fact that more than half of CRT's predictions are outliers ($\delta \geq 10$) while for KDE this number is less than 9\%. The last conventional method, phasor imaging, struggles heavily under added noise and sub-optimal modulation frequencies, leading to nearly all the measurements being incorrectly unwrapped. The U-Net, which is also used in Su et al.~\shortcite{su2018deep}, and two comparison deep learning methods produce similar spatially smoother predictions than the classical methods, however often bin entire patches of the image into the wrong wrap count, leading to a marginally higher rate of outliers than KDE while making more than double the number of correct predictions. Our proposed neural unwrapping method more than doubles the rate of correct predictions when compared to Zhang et al.~\shortcite{zhang2019rapid} baseline, with almost half of the pixels in scenes unwrapped correctly. This is visually confirmed by the significantly more spatially consistent outputs, with object borders aligning to the amplitude measurement. The proposed network is thus outperforms all existing methods in GHz frequency unwrapping.

\subsection{Experimental Assessment}
In this section, we validate the proposed experimental system and we validate that the proposed computational imaging approach provides micron-scale depth resolution that can be captured robustly for diverse macroscopic scenes.

\begin{figure}[t]
\vspace{-4mm}
  \centering
  \includegraphics[width=\linewidth]{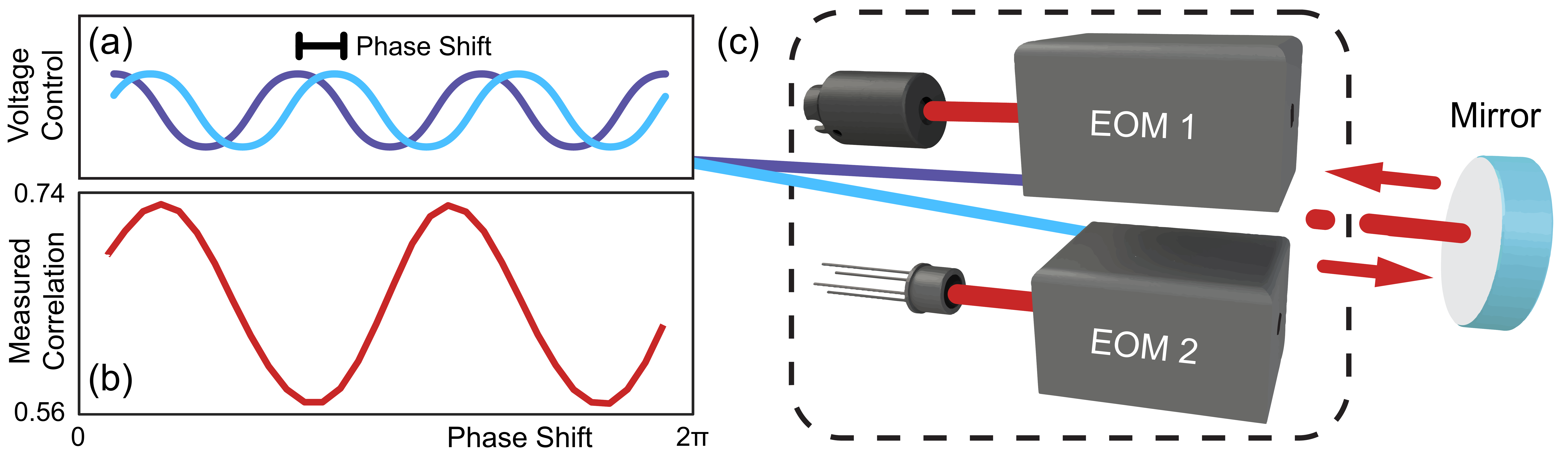}
	  \caption{\label{fig:modulation_validation}%
(c) We validate our optical GHz modulation by capturing the instrument response of the complete system for a fixed mirror with varying phase shifts of the detection EOM, controlled by (a) the voltage-control RF drivers. (b) As predicted in the model, the amplitude measurements accurately follow a sinusoidal function, validating the effective GHz correlation mode.
  }
	\vspace{-3mm}
\end{figure}

\paragraph{Qualitative Reconstruction}
We demonstrate depth captures on diverse real-world scenes as shown in Figure~\ref{fig:experimental_results}.
All scenes were captured with the galvo on the floor plane with respect to the scene, and swept through 16 phase shifts from $0-\pi$, corresponding to 13ms integration time for a single galvo measurement point.
Note that we perform this capture procedure at 532~nm under strong room ambient light, demonstrating the photon efficiency of our system.
 We use a single-frequency ($\omega$) and double-frequency ($2(\omega + 10~MHz)$) pair for a depth measurement.
Figure~\ref{fig:experimental_results} shows that combining the proposed free-space correlation acquisition and neural unwrapping method enables high-fidelity depth reconstruction of all the tested objects with \textit{wide dynamic range}.

\begin{figure}[t]
  \centering
  \vspace{-2mm}
  \includegraphics[width=\linewidth]{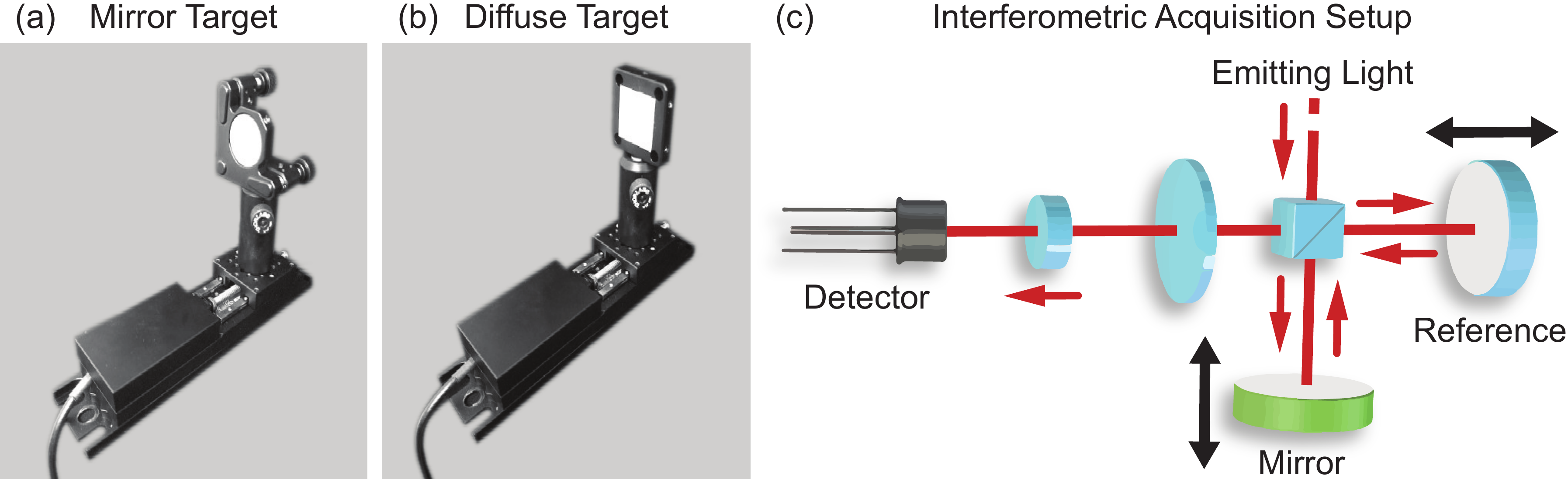}
	  \caption{\label{fig:experimental_precision}%
We measure depth precision of our system for (a) a specular mirror and (b) a diffuse reflector mounted on a linear motion stage, placed at 60~cm distance from the system.
Comparint to the known positions of the target objects in a sweep of the mirror and diffuse reflector, we achieve the depth accuracy of $33.5$\,um with standard deviation of $7.5$\,um, outperforming the post-photoconversion RF-based method and approaching to (c) the optical interferometry.
Note that while interferometry is extremely sensitive to short travel distance and scene reflectance, the proposed method effectively estimates depth  independently of these environmental influences.
\vspace{3mm}
  }
	\resizebox{\columnwidth}{!}{
	\begin{tabular}{ c c c | c  c }
		\toprule
		Method & RMSE Mirror & MAE Mirror & RMSE Diffuse & MAE Diffuse\\
			\midrule
			\midrule
		Interferometry  &20 $\mu$m&20 $\mu$m&14 $\mu$m&14 $\mu$m\\
		RF   &49.5 $\mu$m&48.8 $\mu$m&11800 $\mu$m&11800 $\mu$m\\
		Proposed  &33.5 $\mu$m&32.5 $\mu$m&34.6 $\mu$m&32.9 $\mu$m\\
			\bottomrule
	\end{tabular}
	}
	\caption{\label{table:quantitative_comparison_experimental}%
	Quantitative comparison of the proposed method for diffuse and specular object corresponding to the measurements in Fig.~\ref{fig:experimental_precision}.
	}
	\vspace{-3mm}
\end{figure}
Compared to RF demodulation after photon conversion using the highspeed GaAs 12GHz photodetector as described in Section~\ref{sec:experimental}, the proposed method drastically outperforms post-detection modulation across all experimental tests for the identical photon budget. We tested even for 10$\times$ higher laser power of 30~mW with the same result, again validating the photon-efficiency of the proposed free-space modulation approach.
Our measured phase maps clearly show depth-dependent contours for diverse surface reflectance types (see also Supplemental Material), demonstrating the robustness of the proposed system.
Moreover, our imager handles large reflectance variation of objects from diffuse bust, a highly specular helmet with a very small low diffuse component, and a textured owl object with low albedo components.
We evaluate the impact of our neural phase unwrapping on these challenging scenes compared with the existing KDE~\cite{lawin2016efficient}, recent learned network~\cite{zhang2019rapid} method, and micro ToF phasor unwrapping~\cite{gupta2015phasor} methods.
KDE unwrapping~\cite{lawin2016efficient} struggles with the high frequencies of the proposed system and residual measurements noise, failing to provide high-quality residual measurements.
The other two methods~\cite{gupta2015phasor,lawin2016efficient} also fail to recover correct meaningful geometric structures which can be found in the Supplemental Document. The lookup-table based unwrapping method from~\cite{gupta2015phasor} fails here due to the small modulation bandwidth available in our experimental system. We note that we use the optimal frequency settings for the phasor unwrapping~\cite{gupta2015phasor} in our operating bandwidth.
Our neural phase unwrapping successfully handles high wrapping counts in GHz regime, enabling us to obtain accurate depth maps across all scenes. As such, the experiments validate that the proposed method robustly performs high-frequency correlation depth imaging, outperforming existing approaches and phase unwrapping methods across all tested scenarios. Specifically, the method is robust to surface reflectance, including highly specular objects, such as the helmet, without meaningful diffuse component, and textured object with low-albedo regions, such as the owl.

\begin{figure}[t]
\vspace{-7mm}
  \centering
  \includegraphics[width=\linewidth]{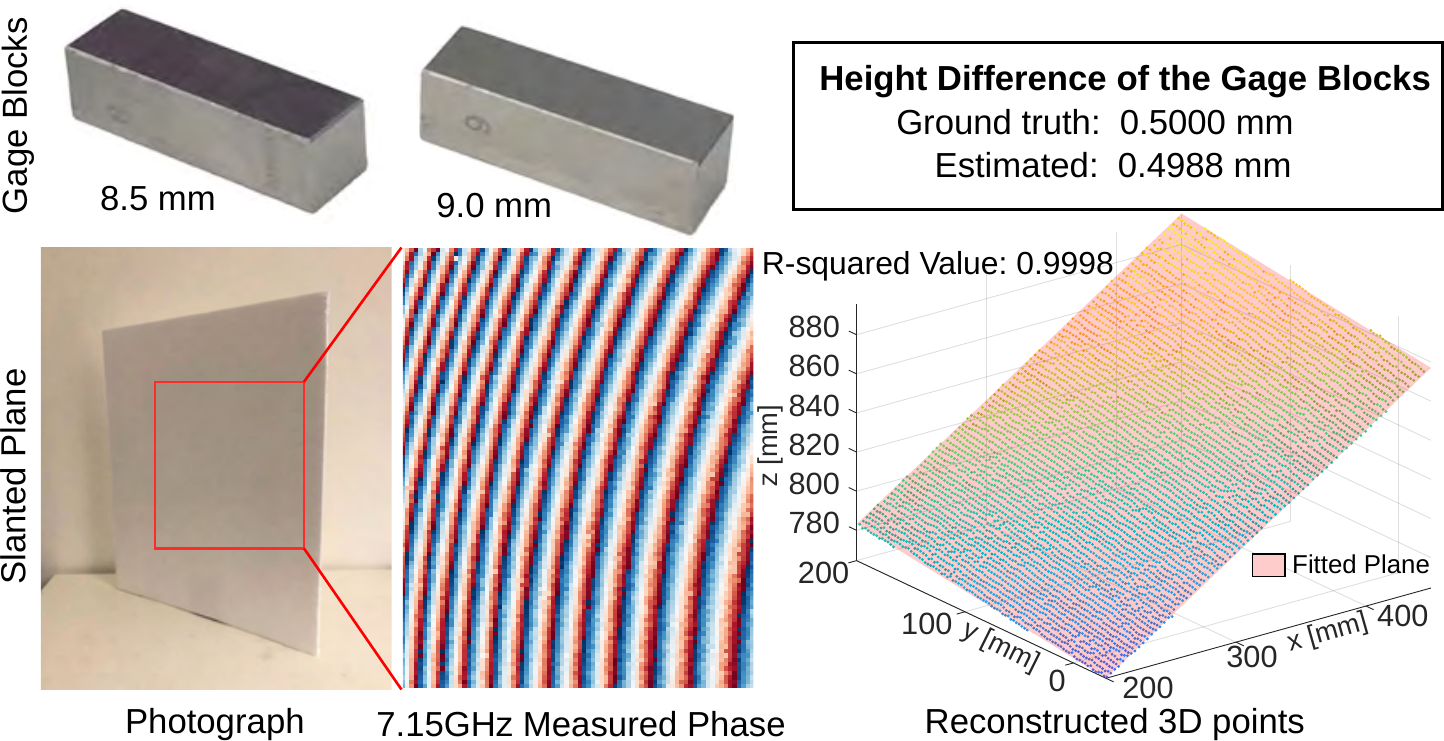}
  \caption{\label{fig:simple_object_eval}%
  {
  As additional illustrative validation of the depth resolution provided by the proposed method we capture gage blocks and a slanted flat plane. Our method accurately recovers micron resolution height differences of the gage blocks. For the plane object, we fit a plane equation to the acquired 3D points and the R-squared value is 0.9998. This validates the effectiveness and precision of our method over larger distances.
  }
  }
	\vspace{-5mm}
\end{figure}
\paragraph{Validation of Correlation Profiles}
We validate the functionality of the proposed imaging system by acquiring correlation measurements as figure of merit. Specifically, we capture measurements of a static target without galvo movement while sweeping the phase of the reference signal driven by the RF driver.
To this end, we place a mirror (Thorlabs PF10-03-P01) at a fixed position and uniformly sample $\psi$ over a range of 0 to 2$\pi$.
Figure~\ref{fig:modulation_validation} confirms that the measured correlation values accurately follow the sinusoidal image formation model from Eq.~\eqref{eq:measured_correlation}.

\paragraph{Quantitative Evaluation of Depth Precision}
We quantitatively evaluate depth precision of our experimental prototype by capturing objects at known distances using a motion stage (Thorlabs MTS50/M-Z8) as shown in Figure~\ref{fig:experimental_precision} and Table~\ref{table:quantitative_comparison_simulation}.
We control the position of the target object that is placed at 60~cm distance from the setup. At this depth offset, we sweep over 1\,mm travel distance with 50\,$\mu$m step size (stage error 0.05 $\mu$m) and estimate corresponding depth values using the proposed method.
Our imaging system obtains micron-scale depth error of 32.5\,um and 32.9\,um for a specular mirror and a diffuse reflector respectively.

{
Furthermore, we measure the height difference between the two metallic precision-fabricated gage blocks at 100~cm distance as shown in Figure~\ref{fig:simple_object_eval}.
The two gage blocks (ACCUSIZE DIN861 Metric, Grade2) are placed on a static mount. The difference of the measured depths, the height difference, is 0.4988\,mm which is only 12\,$\mu$m off from the ground truth 0.5\,mm, demonstrating the precision of our depth acquisition.
Finally, we captured the shape of a large diffuse flat plane at a slanted angle. Once we obtain the depth map from our measurements, we fit a plane equation and the fitting R-squared value is 0.9998, demonstrating the accuracy of our method over longer travel distance than the translation-stage experiment.
}

{
\paragraph{Comparison with MHz AMCW ToF Imaging}
The proposed method perform all-optical GHz modulation for high-resolution depth imaging.
Figure~\ref{fig:ghz_mhz_comparisons} compares our method with the conventional MHz AMCW ToF imaging used in LUCID Helios Flex camera equipped with four VCSEL diodes of cumulatively 8\,mW illumination module at 850\,nm wavelength and 8\,ms exposure which is comparable to the effective photon budget of our system, although less susceptible to ambient light due to the wavelength filter.
Our estimated depth contains geometric details of challenging scenes at correct depth scales, whereas the MHz AMCW ToF suffers from low-precision depth with mm errors on diffuse highly reflecting surfaces and larger cm to 10 cm errors for surface areas of low reflectance. We ignore errors due to multipath effects (naturally suppressed in the proposed scanned method) in this evaluation by focusing on convex object shapes.
In particular, the MHz AMCW ToF fails for recovering correct geometry of bright and the dark spots on the owl, resulting in 200\,mm depth difference, commonly known as texture-dependent depth errors. The method also fails for the faint diffuse component returned for the highly specular helmet scene.

\begin{figure}[t]
\vspace{-3mm}
  \centering
  \includegraphics[width=\linewidth]{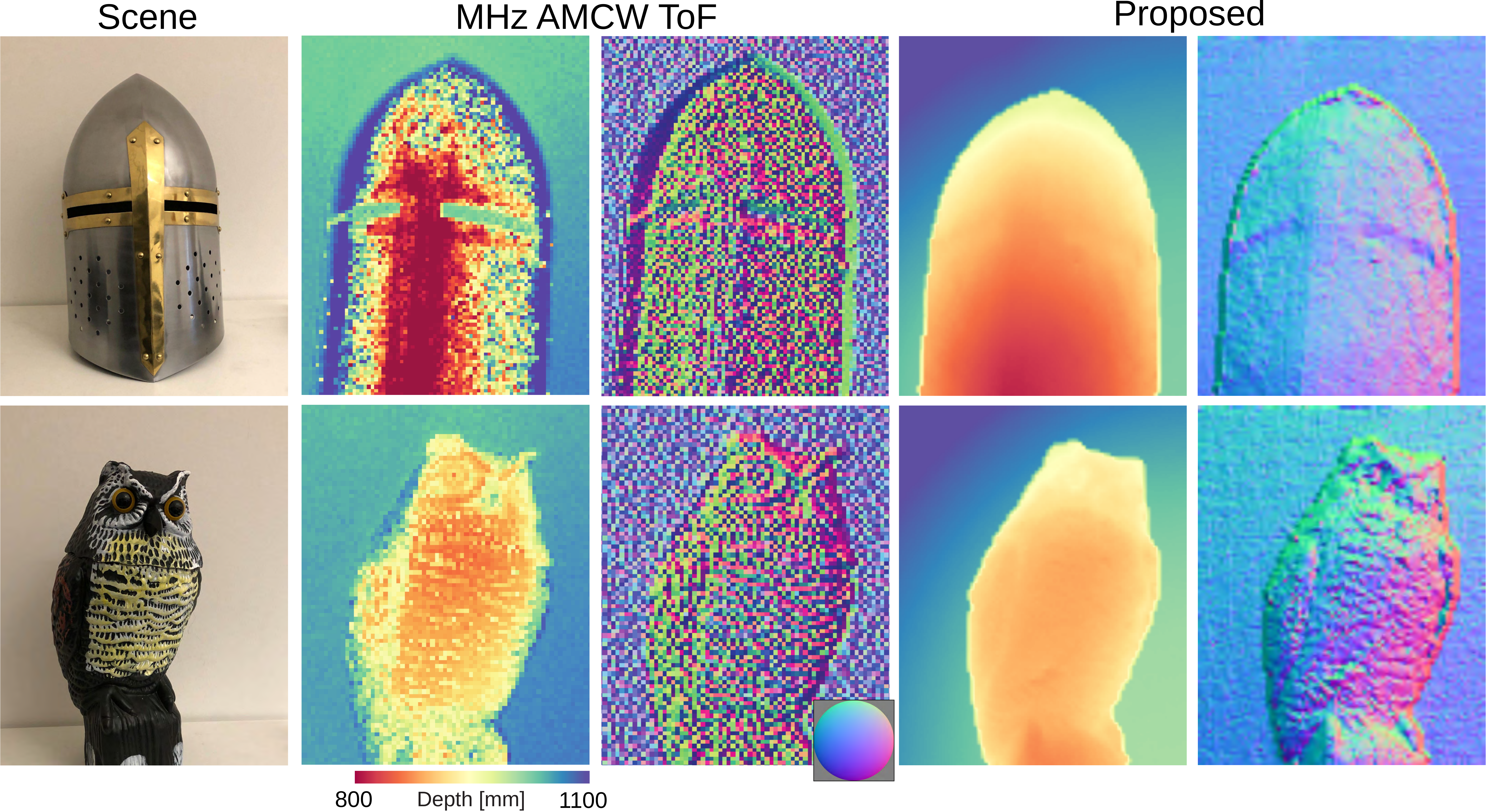}
  \caption{\label{fig:ghz_mhz_comparisons}%
  {Our all-optical GHz ToF imaging captures fine geometric details which cannot be revealed in MHz AMCW ToF imaging.
   In the MHz regime, fundamentally limiting depth resolution and phase contrast by modulation in the analog domain instead of the optical domain, conventional AMCW ToF fails to recover correct depth and fine-grained normals. Noticeably, we observe artifacts on the specular surface of the helmet, which only returns a very faint diffuse component, and the texture-dependent artifacts on the bright and dark spots of the owl object.
  }
  }
	\vspace{-3mm}
\end{figure}
}
\paragraph{Comparison to RF Demodulation and Optical Interferometry}
We compare the proposed method to RF demodulation after photo-conversion and to interferometric depth estimation. To compare to RF demodulation, we use the same highspeed GaAs 12GHz photodetector as above. We note that this was the fastest photodiode available to us, see again Sec.~\ref{sec:experimental} and the Supplement for additional details. To compare interferometric depth estimation with the proposed method, we add a moving reference mirror and an intensity detector so that interference can be detected with the superposed reference and scene beams as shown in Figure~\ref{fig:experimental_precision}(c). To implement this approach with the same proposed setup, we place a beam block in front of the reference mirror when we use the system in the proposed correlation mode. For fair comparison, we unwrap the interferometric data with sequential unwrapping which adds the smallest multiple of 2pi whenever the phase exceeds 2pi.

Figure~\ref{fig:experimental_precision} shows that for a 1~mm sweep at 0.6~m distance, our proposed method with an emitter-decoder setup outperforms the RF demodulation in depth accuracy. Note, that the target is mounted at 60~cm from setup, which demonstrates \emph{micron-scale accuracy at meter-scale range} rather than millimeter-scale range. Our proposed method has a lower depth error than the RF method for RMSE and MAE for both a specular reflector shown in (a), and a diffuse reflector shown in (b). The depth estimates for all methods are shown for a 1 mm range compared to a ground truth for both specular and diffuse reflectors. The RMSE of each method is plotted together in graphs for specular and diffuse reflectors. We validate that, while post-photoconversion performs well for high flux levels, typical diffuse scenes results in low photon counts that are challenging to sense at high frequencies. As such, the RF demodulation approach \emph{fails} for the diffuse scene object. We note that in contrast to direct fast sampling at rates higher than 10~GHz in the RF setup, our \emph{all-optical sensing enables us to get away low-frequency kHz sampling (six orders of magnitude slower) with high SNR}.

The RSME and MAE for the interferometry, RF and proposed methods for specular and diffuse reflectors are shown in Tab.~\ref{table:quantitative_comparison_experimental}. The interferometric depth estimation performs the best in terms of RMSE and MAE for specular and diffuse reflectors as expected. The high accuracy of our method at around 30 microns validates its micron level depth resolution. We note that optical interferometry setup is extremely sensitive to scene scale, system vibrations, to the point where measurements had to be completed remotely from outside the lab and repeated multiple times due to tiny measurement fluctuations. We validate that the proposed method closes the gap to interferometric depth estimation approaches.

\section{Conclusion}
\label{sec:conclusion}
We propose a computational ToF imaging method that correlates light all-optically at centimeter-wave frequencies, without fiber coupling or photon-conversion -- enabling high SNR sensing with more than 10~GHz modulation frequency. To this end, solve two technical challenges: modulating without large signal losses at GHz rates, and unwrapping phase at theses rates which render conventional phase unwrapping methods ineffective. Specifically, we propose polarimetric modulation to optically compute correlation signals in freespace, which we double by passing through the same modulator twice. We combine this with a neural phase unwrapping method to handle high wrapping counts in GHz-frequency measurements, on the order of hundred wraps. The resulting imaging method achieves micron-scale depth reconstruction for macroscopic scenes, robust to materials of low reflectance, highly-specular materials, and ambient light. We experimentally validate that  proposed approach that achieves 30\,um depth resolution, corresponding to < 100 femtosecond temporal resolution, and we demonstrate accurate depth reconstructions, outperforming existing phase-unwrapping and post-photo-conversion ToF methods for \emph{all} synthetic and real-world experiments. As such, we validate that the method fills the gap between interferometric and AMCW ToF. In the future, we envision that the proposed modulation approach could serve as a building block for a diverse array of imaging methods, and as an alternative to femtosecond-scale pulsed imaging could bring an order of magnitude improvement to computational imaging problems that today rely on pulsed direct acquisition, including transient imaging, non-line-of-sight imaging, imaging in scattering media, with the potential for fueling photon-efficient imaging of ultrafast phenomena across disciplines.

\bibliographystyle{ACM-Reference-Format}
\bibliography{bib}

\end{document}